\pdfoutput=1

\documentclass[11pt]{article}

\usepackage{acl}

\usepackage{times}
\usepackage{latexsym}

\usepackage[T1]{fontenc}

\usepackage[utf8]{inputenc}

\usepackage{microtype}

\usepackage[nolist]{acronym} 
\usepackage{booktabs}
\usepackage{subcaption}
\usepackage{graphicx}
\usepackage{amsmath}
\usepackage{bbm}
\usepackage{multirow}
\usepackage{hyperref}
\usepackage{makecell}
\usepackage[switch]{lineno}
\usepackage{soul,sidecap}
\usepackage{romannum}
\usepackage{enumitem}
\usepackage{bbm}

\title{
Consistency Analysis of ChatGPT}

\author{
Myeongjun Erik Jang$^1$~~~
Thomas Lukasiewicz$^{2,1}$~~~
\smallskip 
\\
$^1$\,Department of Computer Science, University of Oxford, UK \\
$^2$\,Institute of Logic and Computation, Vienna University of Technology, Austria \\
myeongjun.jang@cs.ox.ac.uk, thomas.lukasiewicz@tuwien.ac.at\\
}

\begin{document}
\pagenumbering{arabic}

\maketitle
\begin{abstract}
ChatGPT has gained a huge popularity since its introduction. Its positive aspects have been reported through many media platforms, and some analyses even showed that ChatGPT achieved a decent grade in professional exams, adding extra support to the claim that AI can now assist and even replace humans in industrial fields. Others, however, doubt its reliability and trustworthiness. This paper investigates the trustworthiness of ChatGPT and GPT-4 regarding logically consistent behaviour, focusing specifically on semantic consistency and the properties of negation, symmetric, and transitive consistency. Our findings suggest that while both models appear to show an enhanced language understanding and reasoning ability, they still frequently fall short of generating logically consistent predictions. We also ascertain via experiments that prompt designing, few-shot learning and employing larger \acp{LLM} are unlikely to be the ultimate solution to resolve the inconsistency issue of LLMs.
\end{abstract}

\section{Introduction}
AI systems can be more reliable and trustworthy, provided that they behave similarly to humans \cite{de2016almost, jung2019neural}. In this regard, ChatGPT, a \ac{LLM} that simulates human-like conversations~\cite{fares.chatgpt}, is gaining widespread popularity, reaching 100 million users only two months after its launch~\cite{milmo.chatgpt_100m}. In addition to many convenient functions that it provides, ChatGPT has performed astoundingly well on various professional examination cases, including passing the United States Medical Licensing Examination~\cite{kung2023performance}, achieving passing grades in four real exams at the University of Minnesota Law School~\cite{choi2023chatgpt}, and providing decent answers to Operation Management exam questions, which is a core MBA course~\cite{terwiesch2023would}. These surprising results make people believe that \acp{LLM} can assist humans even in professional areas and greatly influence diverse academic and industrial fields.

Others, however, question ChatGPT's reliability, pointing out its overconfidence in generating factually incorrect information~\cite{skopeliti.chatgpt_incorrect_facts}, the inability to comprehend the complexity of human language~\cite{bogost.chatgpt_incorrect}, and imperfect mathematical abilities~\cite{frieder.mathmathics}. Even though these mistakes may appear insignificant in normal daily tasks, they can provoke crucial concerns in conservative and risk-sensitive domains, such as law, medicine, and finance.

A correct behaviour is a crucial aspect in deciding models' trustworthiness by improving the certification\footnote{Trustworthiness = Explanation + Certification \cite{huang2020survey}.} process~\cite{jang2022becel}. In this regard, we mainly investigate the trustworthiness of ChatGPT in terms of logically consistent behaviour. By using the BECEL dataset~\cite{jang2022becel}, which is designed to ascertain whether language models satisfy various types of consistency, we analyse ChatGPT's ability to generate logically consistent predictions based on  the four properties below:

\begin{itemize}[nosep]
    \item Semantic equivalence: $f(X)=f(Y)$ if $X$ and $Y$ mean the same.

    \item Negation property: $f(X) \neq f(\lnot X)$.

    \item Symmetric property: $f(X,Y) \,{=}\, f(Y,X)$.

    \item Transitive property: $X \rightarrow Y \wedge Y \rightarrow Z$ then $X \rightarrow Z$.
\end{itemize}

Our findings suggest that, similarly to previous \acp{PLM}, ChatGPT is also prone to violate logical consistencies. Furthermore, our results ascertain that employing different prompt designs, few-shot learning, and utilising larger \acp{LLM} trained with more data, such as GPT-4, does not necessarily lead to noteworthy enhancements in consistency. Our contributions can be briefly summarised as follows:

\begin{enumerate}[nosep]
    \item We analyse the consistency behaviour of ChatGPT by measuring semantic, negation, symmetric, and transitive consistency.

    \item We observe that ChatGPT achieves a certain level of improvements in negation and transitive consistency compared to previous \acp{PLM}.
    

    
    \item We ascertain that ChatGPT and GPT-4 generate different predictions on text inputs conveying the same meaning.

    \item We confirm that ChatGPT and GPT-4 are both self-contradic\-tory: they violate semantic consistency for paraphrased inputs generated by themselves.

    \item We find that ChatGPT easily violates symmetric consistency, being sensitive to the input sentence order for order-invariant tasks.

    \item Our experiments indicate that prompt design, few-shot learning, and the training of larger \acp{LLM} like GPT-4 are unlikely to be a fundamental and ultimate solution for improving the model's consistency.
    
\end{enumerate}

\section{Related Works}
The consistency of language models has been an important topic in \ac{NLP} but conducted under various definitions. The idea of \textit{semantic consistency} is the most widely used concept in consistency analysis, meaning that a model should make consistent decisions in semantically equivalent contexts~\cite{elazar2021erratum}. Semantic consistency is an indispensable property that should be satisfied in every textual data and \ac{NLP} task. \citet{ravichander2020systematicity} observed that \acp{PLM} are likely to generate different masked language modelling predictions when an object in queries is replaced with its plural form. \citet{elazar2021erratum}, on the other hand, found that \acp{PLM} generate different masked language modelling predictions when given paraphrased queries. \citet{raj2022measuring} proposed an enhanced framework that facilitates the assessment of semantic consistency of \ac{NLG} outputs by introducing agreement functions. Another line of work employed the idea by introducing a consistency regularisation term for training, which penalises the violation of semantic consistency, to train more robust \ac{NLP} models~\cite{wang2021unsupervised, zheng2021consistency, kim2021learn}.

\textit{Symmetric consistency} is a consistency type based on symmetric inference, defined as $f(x,y) = f(y,x)$. This implies that a model should be input-order invariant for tasks where the symmetric property holds. Regarding the \ac{NLI} task, \citet{wang2019if} believed that symmetric consistency applies to data points with ``Not Entailment'', i.e., ``Contradiction'' and ``Neutral'', as a label. They showed that many deep-learning-based \ac{NLI} models change their predictions when the premise and hypothesis are switched. On the other hand, \citet{li2019logic} only considered ``contradiction'' labels for their analysis and ascertained that \ac{NLI} models based on BERT~\cite{BERT} are likely to violate symmetric consistency. \citet{kumar-joshi-2022-striking} performed a symmetric consistency analysis on \ac{NLI} and \ac{STS} tasks in a more conservative manner, arguing that a model should generate not only the same predictions but also the same confidence scores if it is truly input-order invariant. They also observed that \ac{PLM}-based models violated symmetric consistency and introduced a consistency regularisation term to compensate for the issue.

The fundamental idea lying in \textit{negation consistency} is the logical negation property ($p$ is true $\Leftrightarrow$  $\lnot p$ is false \cite{aina2018distributional}). Intuitively, the main idea is that a model's prediction should differ for text inputs delivering the opposite meaning. Several studies investigated the negation consistency of BERT and found that the model often generates the same outputs when asked negated and non-negated masked queries, e.g., ``Birds can lay [MASK]'' and ``Birds cannot lay [MASK]''~\cite{kassner2020negated, ettinger2020bert}. \citet{hossain2020analysis} created negated versions of \ac{NLI} datasets and also observed the violation of negation consistency, suggesting that \acp{PLM} lack the understanding of negation expressions. To alleviate the issue, several works adopted data augmentation to train a model with abundant data containing negation expressions~\cite{asai2020logic, hosseini2021understanding}. \citet{jang2022beyond} introduced the \textit{meaning-matching} task to enhance \acp{PLM}' textual understanding ability and observed performance improvements.


\textit{Transitive consistency} is a consistency type that can measure the deductive reasoning ability. It is derived from transitive inference, represented as $X \rightarrow Y \wedge Y \rightarrow Z$ then $X \rightarrow Z$ for three predicates X, Y, and Z~\cite{gazes2012cognitive, asai2020logic}. In the \ac{NLI} task, \citet{li2019logic} employed the concept to generate four transitive inference rules. For three sentences $P$, $H$, and $Z$, the rules are defined as:

\vspace{-4ex}
\begin{align}
    & E(P,H) \wedge E(H,Z) \rightarrow E(P,Z), \\
    & E(P,H) \wedge C(H,Z) \rightarrow C(P,Z), \\
    & N(P,H) \wedge E(H,Z) \rightarrow \neg C(P,Z), \\
    & N(P,H) \wedge C(H,Z) \rightarrow \neg E(P,Z),
\end{align}
where $E$, $N$, and $C$ refer to entailment, neutral, and contradiction. Based on the rules, they collected a new evaluation set to assess the transitive consistency of BERT-based \ac{NLI} models and showed the inconsistency of the models. Other studies investigated the transitive consistency in \ac{QA}~\cite{asai2020logic, mitchell2022enhancing} and WordNet word senses~\cite{lin-ng-2022-bert} and ascertained that \acp{PLM} lack the ability to perform transitive inference.

\citet{jang2022becel} proposed a universal definition of the language model's consistency and a taxonomy of various consistency types. They also created a new benchmark dataset that enables the evaluation of multiple types of consistencies on various downstream tasks. They assessed diverse \acp{PLM} on the new benchmark and confirmed that, like studies stated above, none of \acp{PLM} show consistent behaviour on all test cases. All the aforementioned works investigated the consistency of \acp{PLM} that emerged before the advent of \acp{LLM} like ChatGPT. To our knowledge, this paper is the first evaluation of \acp{LLM} from various consistency viewpoints.

\section{Experimental Design}

\subsection{Evaluation Scope}
The BECEL dataset provides 19 test sets for assessing five types of consistency on seven downstream tasks. However, we reduced the scope of our experiments mainly because of the competitive usage of OpenAI's \ac{LLM} API. Specifically, our experiments do not consider the additive consistency, as most \acp{PLM} were highly consistent with the additive consistency~\cite{jang2022becel}. Factual consistency, which is a popular subject in \ac{NLP}, is also excluded in our evaluation, as (1)~BECEL  does not contain test cases for a factual inconsistency analysis, and (2) the factual inconsistency of \acp{LLM} is already gaining large attention from many studies~\cite{tam2022evaluating, zhao2023can}, so we decided to focus on exploring relatively less investigated consistency types. As for  downstream tasks, we used the SNLI~\cite{SNLI}, RTE~\cite{RTE}, MRPC~\cite{MRPC}, and WiC~\cite{WiC} datasets. The SST2 and AG-News datasets were not included, as they only contain test cases for evaluating the semantic consistency. Table~\ref{table.data_size_table} shows the size of the test sets for each downstream task and 
consistency type.

\begin{figure*}[t]
	\centering
	\begin{subfigure}[b]{0.4\textwidth}
		\includegraphics[width=\linewidth]{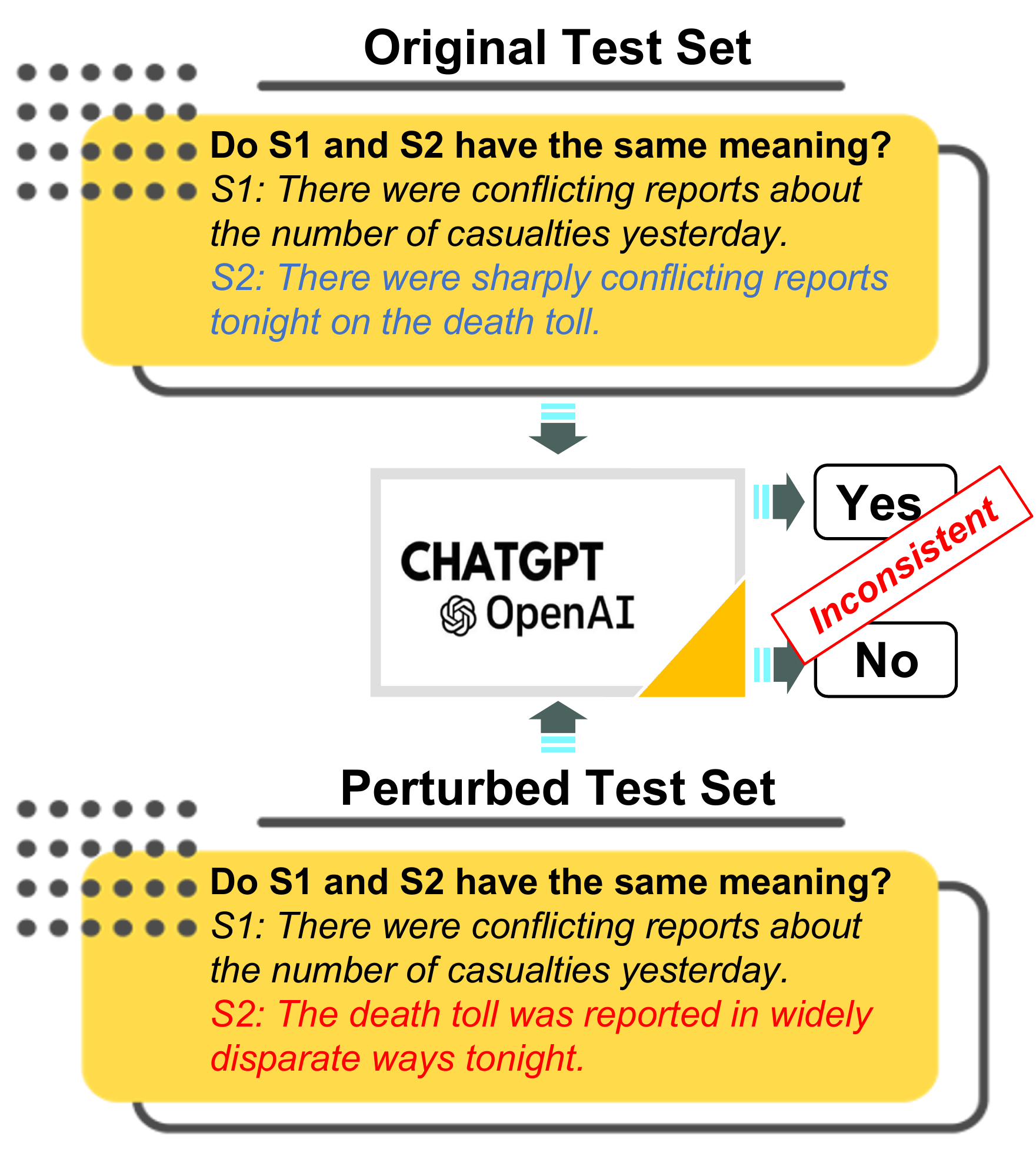}
		\vspace*{-3ex}
		\caption{Semantic Consistency} \label{fig:sem}
\vspace*{1ex}		
 \end{subfigure}
	\begin{subfigure}[b]{0.4\textwidth}
		\includegraphics[width=\linewidth]{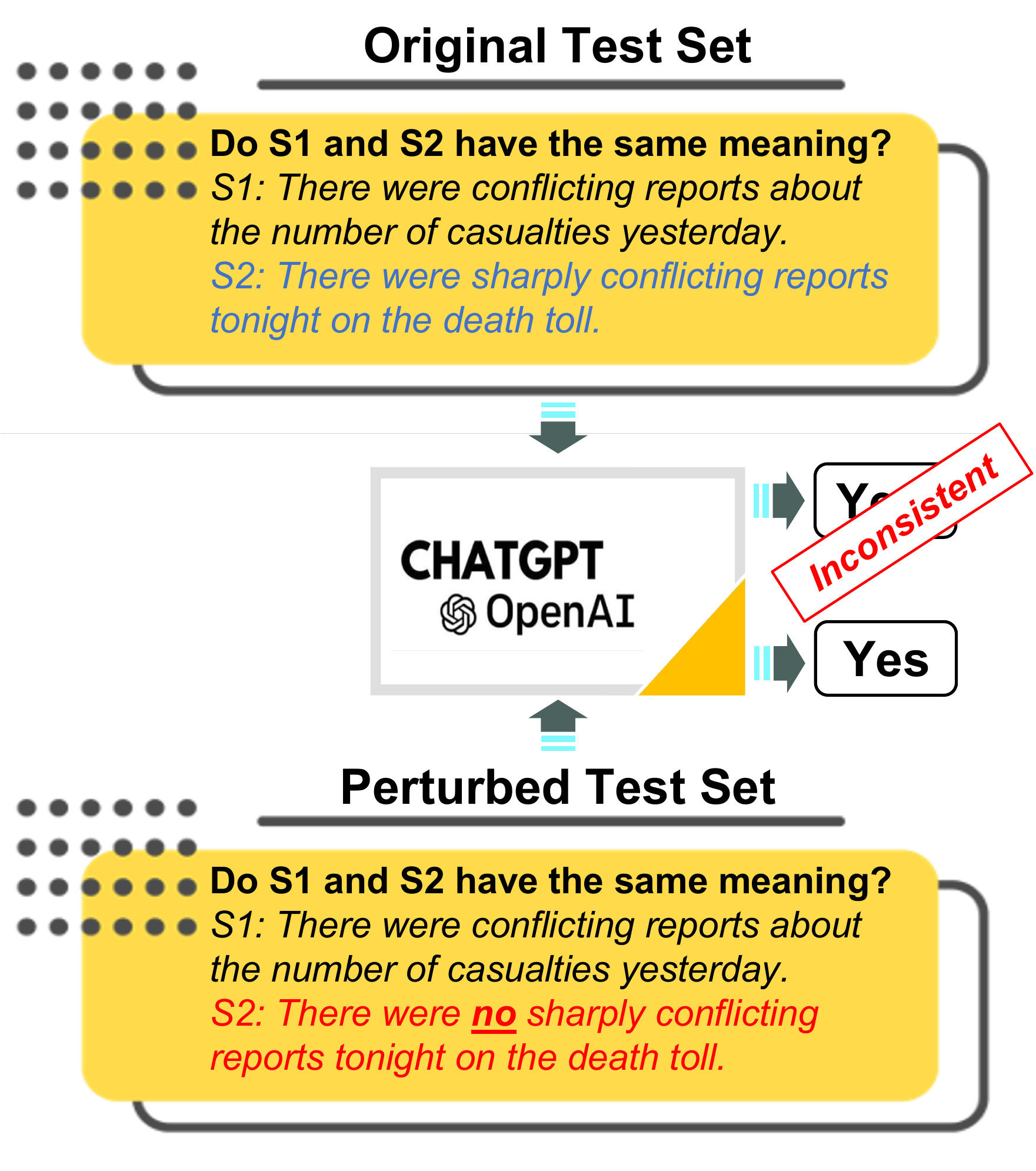}
		\vspace*{-3ex}
		\caption{Negation Consistency} \label{fig:neg}
\vspace*{1ex}	
 \end{subfigure}

	\begin{subfigure}[b]{0.4\textwidth}
		\includegraphics[width=\linewidth]{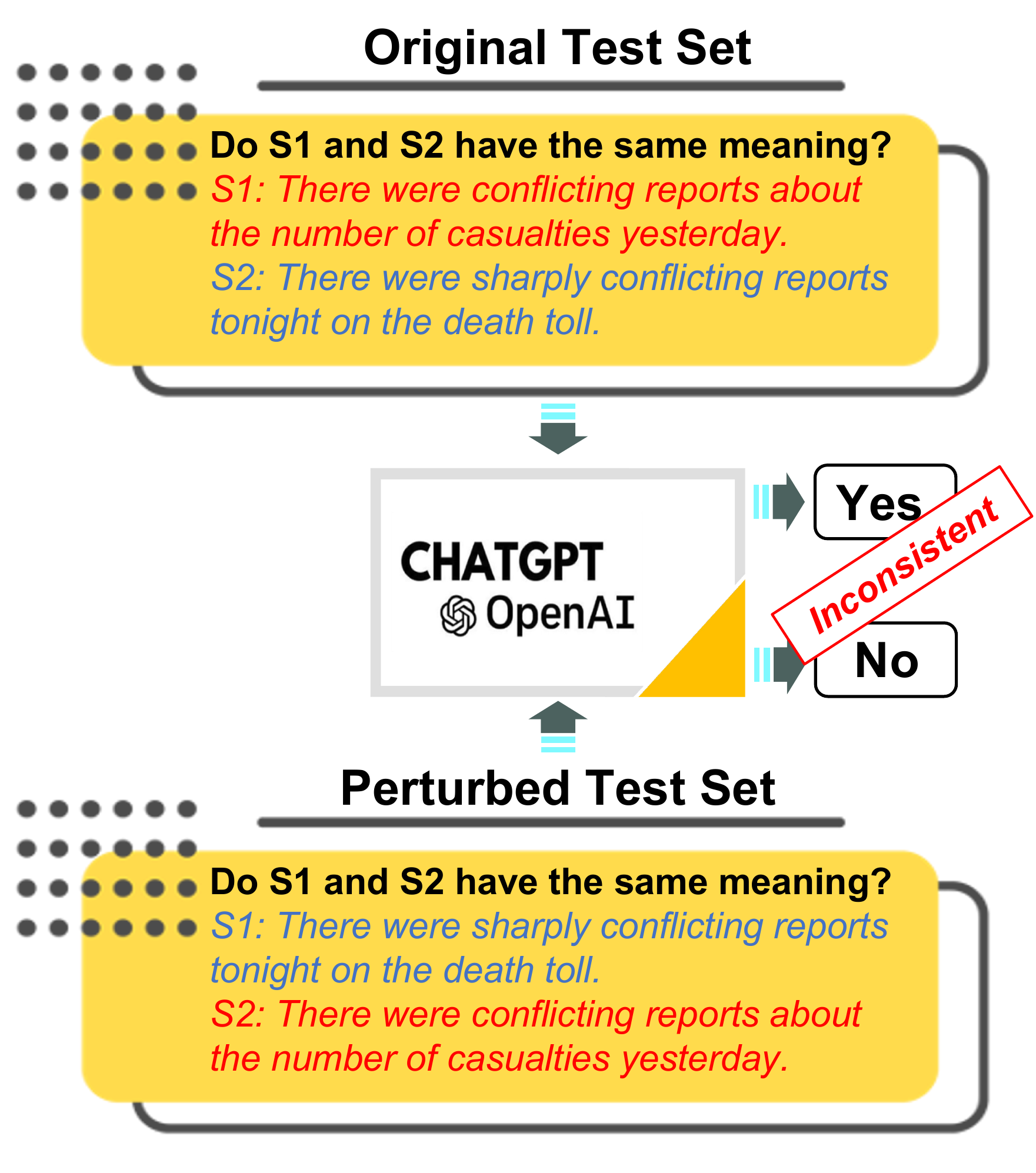}
		\vspace*{-3ex}
		\caption{Symmetric Consistency} \label{fig:sym}
	\end{subfigure}
	\begin{subfigure}[b]{0.4\textwidth}
		\includegraphics[width=\linewidth]{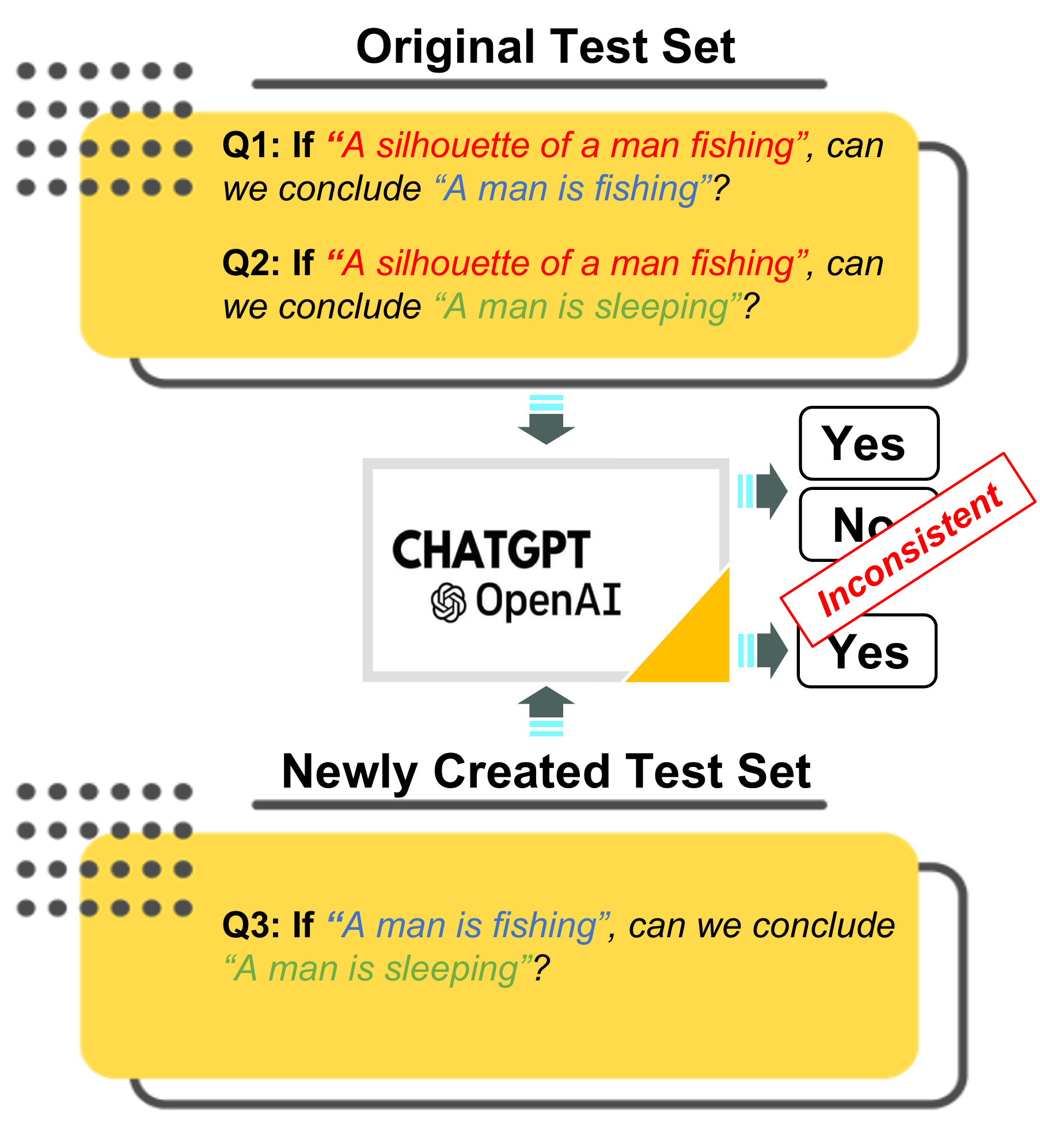}
		\vspace*{-3ex}
		\caption{Transitive Consistency} \label{fig:tran}
	\end{subfigure}
 
	\caption{Consistency evaluation process of (a) semantic, (b) negation, and (c) symmetric consistency on MRPC and (d) transitive consistency on SNLI.}
	\label{fig:task_illustration}
         \vspace{-1ex}

\end{figure*}

\begin{table}[t]
	\begin{center}
		\renewcommand{\arraystretch}{1.5}
		\footnotesize{
			\centering{\setlength\tabcolsep{1.5pt}
		\begin{tabular}{ccccc}
		\toprule
		   & SNLI & RTE & MRPC & WiC \\ \hline
		semantic  & 4,406 & 248 & 202 & 140 \\
		negation  & 2,204 & 153 & 290 & - \\
		symmetric & 3,237 & 1,241 & 3,668 & 5,428 \\
		transitive & 2,375 & - & - & 3,162 \\
		\bottomrule
		\end{tabular}}}
	\vspace{-1ex}
	\caption{Size of the test sets of consistency evaluation data points of the SNLI, RTE, MRPC, and WiC tasks.} 
	\vspace{-1ex}
	\label{table.data_size_table}%
        \vspace{-1ex}
	\end{center}
\end{table}

\subsection{Consistency Evaluation Method}
This section briefly demonstrates the process of consistency evaluation by using the BECEL dataset. The evaluation consists of two steps. First, the predictions of the original test set and its corresponding perturbed test set are generated. Next, the predictions of the two test sets are compared to measure the consistency. 

For the four downstream tasks in our evaluation scope, \citet{jang2022becel} collected the perturbed test sets for semantic and negation consistency evaluation by modifying ``sentence 2'' for the RTE, MRPC, and WiC tasks and ``hypothesis'' for the SNLI task, i.e., generating paraphrase and the opposite meaning sentences for semantic and negation consistency, respectively. They switched the order of the two input texts for symmetric consistency evaluation and created new instances based on existing data examples for assessing transitive consistency. Figure~\ref{fig:task_illustration} illustrates the overall process for measuring the transitive consistency on the SNLI task and the three remaining consistency types on the MRPC task.

\subsection{Generating Predictions}
For test cases of data size above 1K, we sampled 200 data points, keeping a low traffic of the Open\-AI API. We conducted zero-shot experiments by using two prompt versions designed by Eleuther AI~\footnote{\href{https://github.com/EleutherAI/lm-evaluation-harness}{https://github.com/EleutherAI/lm-evaluation-harness}} and \citet{wei2022finetuned} to observe how consistency changes with different prompt design. The prompts of each downstream task and examples are presented in Tables~\ref{table.prompt_example} and \ref{table.prompt_example_wei} in Appendix~\ref{section:prompt_design}. Our experiments are conducted with the use of the \textit{24th May} version of the OpenAI API for ChatGPT and the \textit{20th July} version for GPT-4.

Typically, \acp{LLM} provided a formatted answer, such as ``Yes/No (Equivalent/Not Equivalent)'' for  MRPC, along with (or without) explanations for the decision. However, we observed a few cases where the output deviated from such structured formats. We manually reviewed and modified such cases to match the desired format.

\subsection{Evaluation Metrics}
We used the same inconsistency metric as in \cite{jang2022becel}. Specifically, the metric measures the ratio of predictions that violate the target consistency type. Thus, semantic and symmetric inconsistency count the number of predictions where \acp{LLM} generate different answers for the original and its corresponding perturbed input. In contrast, negation inconsistency counts the results where the two predictions are the same.

Unlike semantic consistency, which holds unconditionally, negation and symmetric consistencies are conditional properties. For example, negation consistency applies when the label is ``Entailment'' for the \ac{NLI} task and ``Equivalent'' for the \ac{STS} task. Regarding symmetric consistency, it applies unconditionally for the \ac{STS} task, but only to ``Not Entailment'' for the \ac{NLI} task. As the BECEL dataset already reflects these conditions, \citet{jang2022becel} calculated the inconsistency metrics using predictions of all test data points, i.e., the condition is determined based on gold labels. However, this can lead to an incorrect estimation of the language model's consistency,  as they are not perfectly accurate. For example, consider the below example of the MRPC task:

\medskip
\noindent \textbf{S1}: In the evening, he asked for six pepperoni pizzas and two six-packs of soft drinks, which officers delivered.

\noindent \textbf{S2}: In the evening, he asked for six pizzas and soda, which police delivered.

\noindent \textbf{S2-neg}: In the evening, he asked for six pizzas and soda, which police did not deliver.

\medskip

The gold label of the \textbf{S1}-\textbf{S2} pair is ``Equivalent'', so predicting the relation between \textbf{S1}-\textbf{S2-neg} as ``Equivalent'' is a violation of negation consistency. However, if the model believes that the answer of the \textbf{S1}-\textbf{S2} pair is ``Not Equivalent'', then generating ``Not Equivalent'' as an answer of the \textbf{S1}-\textbf{S2-neg} pair is hard to be considered as violating negation consistency, because it is the correct answer. Hence, to mitigate the risk of misestimating \acp{LLM}' consistency capability, we introduce an additional metric called conditioned inconsistency metric, which only considers data points where \acp{LLM} make correct predictions for calculating the metric.

\section{Experimental Results}
We now present our experimental results on the consistency performance of \acp{LLM}. The primary analysis target is the comparison between ChatGPT's performance with Eleuther AI's prompt design and those of fine-tuned \acp{PLM}. The BECEL dataset performances of two \acp{PLM} (Electra-large~\cite{electra} and T5~\cite{T5}) are taken from \citet{jang2022becel}. The performance of the prompt design devised by \citet{wei2022finetuned}, GPT-4, and few-shot learning (2-shots in our experiments) will be mainly discussed in Section \ref{section.discussion}.

\begin{table*}[t!]
	\begin{center}
		\renewcommand{\arraystretch}{1.1}
		\footnotesize{
			\centering{\setlength\tabcolsep{5.0pt}
		\begin{tabular}{c|cccccccc|cccccc}
		\toprule
  
        \multicolumn{1}{c|}{\multirow{3}{*}{Model}} & \multicolumn{8}{c|}{Semantic} & \multicolumn{6}{c}{Negation} \\
        & \multicolumn{2}{c}{MRPC} &  \multicolumn{2}{c}{RTE} & \multicolumn{2}{c}{SNLI} & \multicolumn{2}{c|}{WiC} &  \multicolumn{2}{c}{MRPC} &  \multicolumn{2}{c}{RTE} & \multicolumn{2}{c}{SNLI} \\ 
	
        & $\tau_{B}$ & $\tau_{S}$ & $\tau_{B}$ & $\tau_{S}$ & $\tau_{B}$ & $\tau_{S}$ & $\tau_{B}$ & $\tau_{S}$ & $\tau$ & $\tau_{C}$ & $\tau$ & $\tau_{C}$ & $\tau$ & $\tau_{C}$  \\ \hline

	ChatGPT (EAI) & 9.4 & 14.4 & 11.3 & 13.7 & 20.0 & 20.5 &        14.3 & 18.4 & 26.0 & 7.0 & 11.1 & 2.1 & 13.0 & 1.3\\ 
 	ChatGPT (Wei) & 10.9 & 11.4 & 15.3 & 17.3 & 16.5 & 16.5 & 14.3      &       17.2 & \textbf{23.0} & 6.7 & 15.0 & 5.2 & 11.0 & 3.4 \\ 
  	ChatGPT (EAI, 2s) & 17.2 & 13.8 & 10.5 & 10.5 & 17.5 & 14.0 & 30.3 & 34.5 & 42.0 & 4.3 & 15.7 & 13.7 & 23.0 & 20.0 \\ 
  \hline
        GPT-4 (EAI) &  16.3 & 13.4 & 12.5 & 12.5 & 19.0 & 22.0 & 12.9 & 	13.8 & 33.5 & 7.6 & \textbf{7.8} & 6.8 & 8.5 & 1.7 \\ 
        GPT-4 (Wei) & 11.9 & 10.4 & 9.7 & 10.5 & 23.5 & 25.5 & 6.4 & 7.3 & 26.5 & 9.8 & 12.4 & 6.3 & \textbf{4.5} & 0.5 \\  \hline
  
           
Electra-large & 5.5 & - & 8.9 & - & \textbf{7.9} & - & 8.9 & - & 77.0 & - & 17.3 & - & 5.4 & - \\
T5-large & \textbf{4.5} & - & \textbf{8.6} & - & 9.3 & - & \textbf{8.6} & - & 25.2 & - & 15.9 & - & 5.8 & -   \\	\bottomrule
		\end{tabular}}}
	\end{center}
	\vspace{-1ex}
	\caption{Experimental results of the semantic and negation consistency evaluation. ``2s'' refers to two-shot learning. $\tau_{B}$ and $\tau_{S}$ denote the inconsistency of the BECEL dataset and paraphrases generated by \acp{LLM}, respectively. $\tau$ and $\tau_{C}$ refer to the original and conditioned negation inconsistency, respectively. The best performance is in bold. }\label{table.sem_negconsistency}
	\vspace{-1ex}
\end{table*}

\subsection{Semantic Consistency}
\begin{figure}[t!]
	\centering
	\begin{subfigure}[b]{0.48\textwidth}
		\includegraphics[width=\linewidth]{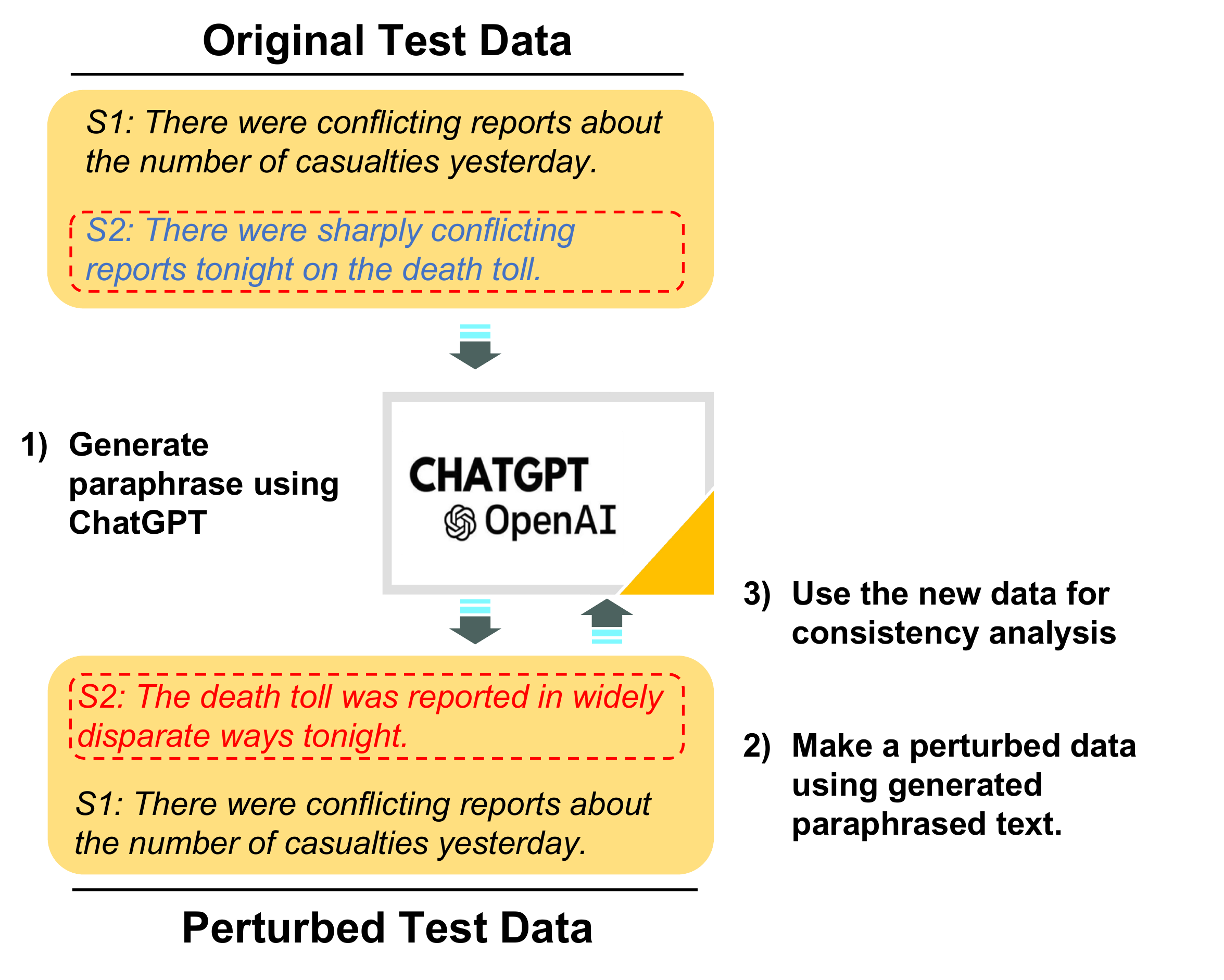}
	\end{subfigure} 
 	\caption{Overall process of measuring semantic consistency by using paraphrases generated by ChatGPT.}\label{figure.selfcontradictory}
    \vspace{-1ex}
\end{figure}

It is widely known that ChatGPT can perform various \ac{NLP} tasks, including summarisation, question answering, and paraphrasing. Therefore, in addition to the original BECEL dataset, we generated paraphrased sentences using ChatGPT and GPT-4 and used them for evaluation. For the WiC task, paraphrased instances were removed if they did not contain the target word. The overall procedure of this evaluation is illustrated in Figure~\ref{figure.selfcontradictory}.


  



The results are summarised in the second column of Table~\ref{table.sem_negconsistency}. In the SNLI task using Eleuther AI's prompt, \acp{LLM} often fail to distinguish ``Neutral'' and ``Contradiction'' classes, potentially leading to an underestimation of consistency. To address this, we integrate the two classes into a unified ``Not Entailment'' class specifically for this scenario. Our experimental results show that ChatGPT produces much higher levels of inconsistency in the BECEL dataset than fine-tuned \acp{PLM}, suggesting ChatGPT's limited capacity for making logically consistent predictions. Moreover, we ascertain that ChatGPT is self-contradictory, i.e., it produces inconsistent outputs for self-generated paraphrase sentences with a probability exceeding  10\%. This implies that ChatGPT failed to generate a proper paraphrased sentence or to capture the meaning of texts delivering the same meaning; either case undermines its reliability. Several examples where ChatGPT violates semantic consistency are presented in Table~\ref{table.semantic_example}. More examples are available in Table~\ref{table.more_semantic_example} in Appendix~\ref{section.example}.






\begin{table*}[ht]
	\begin{center}
		\renewcommand{\arraystretch}{1.0}
		\footnotesize{
			\centering{\setlength\tabcolsep{3pt}
    		      \begin{tabular}{p{0.48\linewidth} | p{0.48\linewidth}}
					\toprule
                    
                    \multicolumn{2}{c}{\makecell{\textsc{Task}: SNLI,  \textsc{Paraphrase Type}: BECEL}} \\
                    
                    \makecell[c]{\textsc{Original Inputs}} & \makecell[c]{\textsc{Perturbed Inputs}} \\
                    
                    \textsc{Premise}: Kids play in the water in the middle of the street. & 
                    \textsc{Premise}: Kids play in the water in the middle of the street. \\
                    
                    \textsc{Hypothesis}: Kids are running from zombies. & 
                    \textsc{Hypothesis}: Children are fleeing from zombies. \\
                    
                    \makecell[l]{\textsc{Prediction}: Not Entailment (Contradiction)} & 
                    \makecell[l]{\textsc{Prediction}: Entailment} \\
                    \midrule

                    \multicolumn{2}{c}{\makecell{\textsc{Task}: MRPC, \textsc{Paraphrase Type}: ChatGPT}} \\
                    
                    \makecell[c]{\textsc{Original Inputs}} & \makecell[c]{\textsc{Perturbed Inputs}} \\
                    
                    \textsc{S1}: Looking to buy the latest Harry Potter ? & 
                    \textsc{S1}: Looking to buy the latest Harry Potter ? \\
                    
                    \textsc{S2}: Harry Potter 's latest wizard trick ? & 
                    \textsc{S2}: The newest magical feat of Harry Potter ? \\
                    
                    \makecell[l]{\textsc{Prediction}: Not Equivalent} & 
                    \makecell[l]{\textsc{Prediction}: Equivalent} \\
                    \midrule

		\end{tabular}}}
	\vspace{-1ex}
	\caption{Examples of semantic consistency violation.} \label{table.semantic_example}%
	\end{center}
    \vspace{-1ex}
\end{table*}

\subsection{Negation Consistency}
The third column of Table~\ref{table.sem_negconsistency} presents the experimental results of the negation consistency evaluation. Compared to the fine-tuned \acp{PLM}, ChatGPT attains a lower negation inconsistency in general. A large improvement has been made in the RTE task and MRPC task compared to the Electra-large model. Also, the conditional inconsistency is extensively small compared to the original inconsistency metric, attaining an average of 4.6\% and almost perfectly consistent on the SNLI task. The difference with the original inconsistency metric substantiates the introduction of the conservative evaluation metric, i.e., conditional inconsistency, aimed at more precise evaluations. The experimental results suggest that ChatGPT can better understand negation expressions and antonyms, which has been a critical issue for \acp{PLM} trained in a self-supervised fashion~\cite{kassner2020negated, ettinger2020bert, hossain2020analysis, hosseini2021understanding, jang2022beyond}. We believe that incorporating human feedback into ChatGPT training~\cite{ouyang2022training} plays a crucial role in learning the meaning of negation expressions and antonyms, compared to previous \acp{PLM} that infer their meaning based on the distributional hypothesis by simply relying on the context information. Investigating the impact of providing human feedback on learning textual meaning is an interesting future research direction. Table~\ref{table.neg_sym_examples} presents an example of negation consistency violation. More examples can be found in Table~\ref{table.more_neg_sym_examples} in Appendix~\ref{section.example}.




  


\subsection{Symmetric Consistency}
The results of the symmetric consistency evaluation are shown in the second column of Table~\ref{table.sym_tran_consistency}. Compared to the best-performing \ac{PLM}, ChatGPT produces three times higher symmetric inconsistency in the MRPC task and five times higher in the RTE task. The conditioned inconsistency was lower than the original inconsistency in the RTE and SNLI tasks, and higher in the WiC task. Although the inconsistency rate for the SNLI task might be considered trivial, it should not be overlooked, considering the simple nature of the symmetric property. Consider a medical-domain model that takes a list of symptoms and generates prescriptions. For such a model, which should operate conservatively, its trustworthiness would be significantly undermined if it were to generate entirely different prescriptions whenever the order of symptoms changes, even if the probability of such occurrence is exceedingly low. Hence, an effort should be made to make \acp{LLM} satisfy logical consistencies to enhance their reliability and safe usage in real-world applications. Table~\ref{table.neg_sym_examples} presents an example of symmetric consistency violations. More examples are presented in Table~\ref{table.more_neg_sym_examples} in Appendix~\ref{section.example}.

\begin{table*}[t!]
	\begin{center}
		\renewcommand{\arraystretch}{1.1}
		\footnotesize{
			\centering{\setlength\tabcolsep{6.0pt}
		\begin{tabular}{c|cccccccc|cc}
		\toprule
		\multicolumn{1}{c|}{\multirow{3}{*}{Model}} & \multicolumn{8}{c|}{Symmetric} & \multicolumn{2}{c}{Transitive} \\
  
        & \multicolumn{2}{c}{MRPC} &  \multicolumn{2}{c}{RTE} & \multicolumn{2}{c}{SNLI} & \multicolumn{2}{c|}{WiC} & \multicolumn{1}{c}{SNLI} & \multicolumn{1}{c}{WiC} \\ 
		& $\tau$ & $\tau_{C}$ & $\tau$ & $\tau_{C}$ & $\tau$ & $\tau_{C}$ & $\tau$ & $\tau_{C}$ & $\tau$ & $\tau$\\ \hline


         ChatGPT (EAI) & 13.0 & - & 45.0 & 19.7 & 5.0 & 1.4 & 20.0 & 25.8 & \textbf{2.1} & 12.3 \\
        ChatGPT (Wei) & 14.5 & - & 45.5 & 33.3 & 6.5 & 3.6 & 23.5 & 24.2 & 2.6 & 14.9 \\ 
        ChatGPT (EAI, 2s) & 31.0 & - & 55.0 & 37.2 & 14.0 & 2.3 & 39.5 & 43.0 & 5.8 & 28.9 \\\hline
        GPT-4 (EAI) & 11.0 & - & 54.5 & 20.7 & \textbf{2.0} & 1.0 & 6.5 & 4.1 & 2.3	& \textbf{3.6} \\ 
	GPT-4 (Wei) & 9.0 & - & 17.0 & 6.9 & 2.5 & 2.0 & 9.5 & 8.1 & 2.9 & 4.3 \\    \hline
 
    Electra-large & 5.3 & - & \textbf{6.7} & - & 6.4 & - & 7.9 & - & 2.5 & 46.5 \\
  T5-large & \textbf{4.2} & - & 8.0 & - & 8.3 & - & \textbf{6.3} & - & 2.9 & 45.3 \\ 

		\bottomrule
		\end{tabular}}}
	\end{center}
	\vspace{-1ex}
	\caption{Experimental results of the symmetric and transitive consistency evaluation. ``2s'' refers to two-shot learning. $\tau$ and $\tau_{C}$ denote the original and conditioned symmetric inconsistency, respectively. The best performance is in bold.}\label{table.sym_tran_consistency}
	\vspace{-1ex}
\end{table*}

\begin{table*}[ht]
	\begin{center}
		\renewcommand{\arraystretch}{1.0}
		\footnotesize{
			\centering{\setlength\tabcolsep{3pt}
        		\begin{tabular}{p{0.48\linewidth} | p{0.48\linewidth}}
					\toprule

                    \multicolumn{2}{c}{\makecell{\textsc{Task}: MRPC, \textsc{Consistency Type}: Negation}} \\
                    
                    \makecell[c]{\textsc{Original Inputs}} & \makecell[c]{\textsc{Perturbed Inputs}} \\
                    
                    \textsc{S1}: He arrives later this week on the first state visit by a US President . & 
                    \textsc{S1}: He arrives later this week on the first state visit by a US President . \\
                    
                    \textsc{S2}: Mr Bush arrives on Tuesday on the first state visit by an American President . & 
                    \textsc{S2}: Mr Bush \underline{doesn't} arrive on Tuesday on the first state visit by an American President. \\
                    
                    \makecell[l]{\textsc{Prediction}: Equivalent} & 
                    \makecell[l]{\textsc{Prediction}: Equivalent} \\
                    \midrule

                    \multicolumn{2}{c}{\makecell{\textsc{Task}: SNLI,  \textsc{Consistency Type}: Symmetric}} \\
                    
                    \makecell[c]{\textsc{Original Inputs}} & \makecell[c]{\textsc{Perturbed Inputs}} \\
                    
                    \textsc{Premise}: There is a man climbing as the boy holds the rope & 
                    \textsc{Premise}: A man holds a rope for a boy who's about to climb a wall. \\
                    
                    \textsc{Hypothesis}: A man holds a rope for a boy who's about to climb a wall. & 
                    \textsc{Hypothesis}: There is a man climbing as the boy holds the rope \\
                    
                    \makecell[l]{\textsc{Prediction}: Not Entailment (Contradiction)} & 
                    \makecell[l]{\textsc{Prediction}: Entailment} \\
                    \midrule

                    \multicolumn{2}{c}{\makecell{\textsc{Task}: WiC,  \textsc{Consistency Type}: Transitive}} \\

                    \multicolumn{2}{c}{\textsc{Original Inputs}} \\
                    \textsc{Sentence1}: You must carry your camping gear. & 
                    \textsc{Sentence1}: The airwaves carry the sound. \\
                                        
                    \textsc{Sentence2}: Sound carries well over water. & 
                    \textsc{Sentence2}: Sound carries well over water. \\
                    \textsc{Word}: carry & \textsc{Word}: carry \\
                    
                    \textsc{Prediction}: Not Equivalent  & 
                    \textsc{Prediction}: Equivalent \\

                    \multicolumn{2}{c}{\textsc{Newly Created Inputs}} \\
                    \multicolumn{2}{l}{\makecell{\textsc{Sentence1}: The airwaves carry the sound. \textsc{Sentence2}: You must carry your camping gear. \textsc{Word}: carry \\                    \textsc{Prediction}: Equivalent
                    }} 
                    \\ \bottomrule
                    
		\end{tabular}}}
	\vspace{-1ex}
	\caption{Examples of negation, symmetric, and transitive consistency violations.} \label{table.neg_sym_examples}%
	\end{center}
    \vspace{-1ex}
\end{table*}

\subsection{Transitive Consistency}
The third column in Table~\ref{table.sym_tran_consistency} presents the transitive consistency evaluation results. In contrast to other consistency types where ChatGPT performed similar or worse than fine-tuned \acp{PLM}, it shows better results than fine-tuned \acp{PLM}, especially with notable improvements in the WiC dataset. It is very interesting that ChatGPT is better at higher-level logical reasoning like transitive inference, but fails in simpler logical properties, such as the symmetric property. The results indicate that combining the advantages of fine-tuned \acp{PLM} and \acp{LLM} can pave the way for developing more logically consistent language models. An example that violates transitive consistency is presented in Table~\ref{table.neg_sym_examples}. More examples are listed in Table~\ref{table.more_transitive_example} in Appendix~\ref{section.example}.

\subsection{ChatGPT's Explainablity}
Providing explanations is a core property of trustworthy systems~\cite{huang2020survey}. It is widely known that generative language models can provide answers with explanations. However, recent studies confirmed that explanations generated by early-stage \acp{LLM} like GPT-3 are often logically inconsistent and unreliable~\cite{jung-etal-2022-maieutic,wang-etal-2023-scott}. We confirmed that the identical problem persists in ChatGPT. We observed several cases that the generated explanations seem plausible but contradictory to other generated explanations with identical context~\cite{camburu2020make}, or not faithful. Table~\ref{table.explanation_example} presents some examples. For the first example, the explanations of the original and perturbed inputs contradict each other. In the second example, the explanation of the perturbed input is not correct, i.e., the input did mention the age and gender of the person pushing the shopping cart (``boy'' and ``A young man''). It would be interesting for future work to quantitatively evaluate the quality of explanations provided by \acp{LLM} by employing inconsistent explanation evaluation frameworks~\cite{camburu2020make, jang2023know}.

\begin{table*}[ht]
	\begin{center}
		\renewcommand{\arraystretch}{1.0}
		\footnotesize{
			\centering{\setlength\tabcolsep{1.5pt}
           		\begin{tabular}{p{0.98\linewidth}}
					\toprule
                    \makecell[c]{\textsc{Task}: MRPC,  \textsc{Consistency Type}: Symmetric} \\

                   \textsc{Original Question}: ``Sentence 1: In the evening, he asked for six pepperoni pizzas and two six-packs of soft drinks, which officers delivered. Sentence 2: In the evening, he asked for six pizzas and soda , which police delivered. Question: Do both sentences mean the same thing? Answer:`` \\
                    
                    \textsc{Original Output}: ``Yes, both sentences mean the same thing. They convey that in the evening, the person made a request for six pepperoni pizzas and two six-packs of soft drinks, and that the officers delivered the items. The second sentence is a simplified version of the first one, where `pizzas` refers to pepperoni pizzas, and `soda` refers to soft drinks.`` \\ \\
                    
                    \textsc{Perturbed Question}: ``Sentence 1: In the evening, he asked for six pizzas and soda , which police delivered. Sentence 2: In the evening, he asked for six pepperoni pizzas and two six-packs of soft drinks, which officers delivered. Question: Do both sentences mean the same thing? Answer:``\\
                    
                    \textsc{Perturbed Output}: ``No, the two sentences do not mean the same thing. The first sentence says that he asked for six pizzas and soda, while the second sentence says that he asked for six pepperoni pizzas and two six-packs of soft drinks. The type of pizza and the specific drinks requested are different in the two sentences.`` \\
                    \midrule

                    \makecell{\textsc{Task}: SNLI, \makecell{Semantic-BECEL}} \\

                    \textsc{Original Question}: ``An enthusiastic boy pushes a shopping card containing a smiling boy. Question: A boy is pushing\\ a shopping cart. True, False or Neither? Answer:`` \\
                    \textsc{Original Output}: ``True`` \\ \\

                    \textsc{Perturbed Question}: ``An enthusiastic boy pushes a shopping card containing a smiling boy. Question: A young man\\ is pushing a shopping cart. True, False or Neither? Answer:`` \\
                    \makecell[l]{\textsc{Perturbed Output}: ``Neither (the description does not mention the age or gender of the person pushing the shopping cart, \\only that the person inside the cart is a smiling boy.)``
} \\ \hline
                    
		\end{tabular}}}
	\vspace{-1ex}
	\caption{Examples of ChatGPT's output with explanations.} \label{table.explanation_example}%
	\end{center}
    \vspace{-1ex}
\end{table*}

\section{Discussion}\label{section.discussion}
\noindent\textbf{Can Prompt Design be a Solution?} Prompts are input text consisting of a task demonstration and, for a few-shot task, some examples~\cite{lester2021power}. Prompt design has been shown to be an effective method of regulating the behaviour of GPT-3~\cite{GPT3}. Hence, one might argue that searching for an optimal prompt for each task can improve consistency. However, our experimental results provide sceptical evidence for this claim. When comparing the performance of ChatGPT using Eleuther AI's prompts and those designed by \citet{wei2022finetuned}, there was no statistically significant difference in performance across all consistency types and downstream tasks at a confidence level of 0.05. In the case of GPT-4, similar results were observed, where a statistically significant difference at a confidence level of 0.05 was found only in symmetric consistency in the RTE task. We believe that a primary cause contributing to this phenomenon is an inherent characteristic of machine learning: inductive reasoning. The underlying idea behind prompt design is that prompts created by experimenters are not optimal, because language models might have acquired target information from completely different contexts~\cite{jiang2020can}. That is, it can resolve the matter of inconsistency if and only if numerous consistency properties are reflected in ChatGPT's inductive bias, which is technically not feasible. Moreover, consistency improvements with prompt design can be considered another violation of semantic consistency,  because the prompts will deliver identical semantic meaning, i.e., task description.

\noindent \textbf{Can Few-shot Learning be a Solution?} It is widely known that providing few-shot examples generally leads to higher accuracy compared to the zero-shot setting. To ascertain whether this principle extends to consistency, we conducted additional few-shot experiments on ChatGPT by providing two-shot examples and using Eleuther AI's prompt. However, the findings suggest that providing few-shot examples is not beneficial to improving consistency. When compared with the result of ChatGPT employing the same prompt design but under a zero-shot setting, we observe statistically significant increases in inconsistencies across the majority of test cases. The most substantial rise in inconsistency values is evident in the context of symmetric consistency and the WiC task. We even identified several instances where the model altered its decision when the order of two-shot examples was switched. These experimental outcomes indicate that employing a few-shot approach does not represent a definitive solution for enhancing the consistency of \acp{LLM}.

\noindent \textbf{Can Increasing Data and Model Size be a Solution?}  Another possible way to improve consistency is training a larger \acp{LLM} with more abundant training data, as larger models generally outperform smaller ones in numerous \acp{NLP} downstream tasks. However, our experimental results reveal the limitations of this approach. Through a comprehensive comparison of ChatGPT and GPT-4 performance employing identical prompt designs, it was ascertained that GPT-4 does not necessarily exhibit superior performance in comparison to ChatGPT from a consistency perspective. While GPT-4 did demonstrate enhanced consistency in several test scenarios, including symmetric and transitive consistency in the WiC task across both prompt designs, symmetric consistency in the RTE task, and negation consistency in the SNLI task using prompts devised by \citet{wei2022finetuned}, no discernible improvements were observed in the remaining test cases, which constitutes a portion of 77\% of among the total test scenarios. In addition, GPT-4 also exhibited a high level of self-contradiction, just like ChatGPT.

Moreover, increasing the size of the data and model is a technically unsustainable strategy. First, the data collection procedure requires tremendous effort and is challenging to cover all possible variations, especially for consistency types that demand abundant linguistic resources, such as semantic consistency. Second, even if we successfully expand the data, it is doubtful whether we can afford to update an \ac{LLM} with each new dataset iteration. Considering the ever-changing information, the data expansion and update of an \ac{LLM} should be performed continuously, as neglecting to do so may raise the concern of outdated information~\cite{zhuo2023red, wen2023future}. However, training an \ac{LLM} entails tremendous financial and environmental costs~\cite{bender2021dangers}. For instance, training a BERT-base model without hyperparameter tuning requires a CO2 emission of 650kg, which is comparable to flying from New York to San Francisco for one passenger~\cite{strubell2019energy}. A simple expectation of CO2 emission for re-training ChatGPT and GPT-4 would amount to 1,033t and 1,0240t, respectively,\footnote{ChatGPT and GPT-4 are approximately 1,590 and 16,000 times larger than BERT-base.} while a human is responsible for 5t CO2 emission per year. The continuous emission of such a substantial volume of greenhouse gases would have a detrimental impact on the environment of modern society facing the global climate crisis.

\section{Summary and Outlook}
The advent of ChatGPT is accelerating the developments in the \ac{NLP} field driven by \acp{LLM}. Its outstanding performance captured considerable attention, resulting in many articles, posts, and analyses highlighting ChatGPT's positive aspects across numerous media. There are others, however, who question its reliability based on the model's faulty behaviours. To this end, this study aims to examine the trustworthiness of ChatGPT in terms of the language model's consistency. 

In this paper, we have investigated the consistency behaviour of ChatGPT across four consistency types and downstream tasks. Our experimental results demonstrate that ChatGPT achieves a certain level of enhanced language understanding ability, especially in negation expressions and antonyms. It also exhibits improved deductive reasoning ability with lower transitive inconsistencies compared to the earlier version of \acp{PLM}. However, while ChatGPT exhibits enhanced negation and transitive consistency, this does not mean that the model is perfectly consistent, i.e., it still makes mistakes that violate the logical properties, and the frequency of such occurrences is non-negligible. Also, contrary to the widespread belief regarding the outstanding performance of ChatGPT, its performance across various consistency types falls short of expectations. It frequently changes its decision when an input text is replaced with a paraphrased sentence, even though it is generated from ChatGPT itself. Moreover, in input-order invariant tasks, ChatGPT is prone to make a different decision when the order of the input sentences is switched. These issues are also observed in GPT-4 or with the use of different prompt designs and few-shot learning, indicating that these approaches are unlikely to be a fundamental remedy. Given how simple and natural the symmetric and semantic consistencies are in human reasoning, violating these consistencies can be a huge blow to \acp{LLM}' trustworthiness. These fallacious behaviours are especially lethal to domains operating conservatively and at high risk. Although \acp{LLM} are a revolutionary technique that brought an unprecedented era to \ac{NLP}, such issues should be resolved before these models are applied in real applications, particularly considering the huge economic and environmental costs consumed for developing \acp{LLM}.

\section*{Limitations}
Limitations of our work include that the data instances of several test cases (e.g., SNLI dataset and symmetric evaluation cases) are sampled due to the heavy usage of ChatGPT. Conducting experiments on whole data points will provide a more precise comparison with baseline models.

We were unable to study the model's consistency on longer documents, e.g., document-level \ac{NLI} task, because there are no publicly available datasets for evaluating consistency on long documents. We leave this as future work, as a consistency analysis of long documents would be very appropriate for ChatGPT studies.

We performed a qualitative evaluation of explanations generated by ChatGPT, but a quantitative analysis was omitted due to the resources and time required for the evaluations. We leave adopting several explanation quality evaluation frameworks~\cite{camburu2020make, jang2023know} to ChatGPT's explanations as future work.


\section*{Acknowledgements}
This work was partially supported by the Alan Turing Institute under the EPSRC grant EP/N510129/1 and by the AXA Research Fund. We also acknowledge the use of Oxford’s ARC facility, of the EPSRC-fun\-ded Tier 2 facility JADE~\Romannum{2} (EP/ T022205/1), and of GPU computing support by Scan Computers International Ltd.

\bibliography{custom}

\begin{thebibliography}{53}
\expandafter\ifx\csname natexlab\endcsname\relax\def\natexlab#1{#1}\fi

\bibitem[{Aina et~al.(2018)Aina, Bernardi, and
  Fern{\'a}ndez}]{aina2018distributional}
Laura Aina, Raffaella Bernardi, and Raquel Fern{\'a}ndez. 2018.
\newblock A distributional study of negated adjectives and antonyms.
\newblock In \emph{CEUR Workshop Proceedings}, volume 2253.

\bibitem[{Asai and Hajishirzi(2020)}]{asai2020logic}
Akari Asai and Hannaneh Hajishirzi. 2020.
\newblock \href {https://doi.org/10.18653/v1/2020.acl-main.499} {Logic-guided
  data augmentation and regularization for consistent question answering}.
\newblock In \emph{Proceedings of the 58th Annual Meeting of the Association
  for Computational Linguistics}, pages 5642--5650, Online. Association for
  Computational Linguistics.

\bibitem[{Bender et~al.(2021)Bender, Gebru, McMillan-Major, and
  Shmitchell}]{bender2021dangers}
Emily~M. Bender, Timnit Gebru, Angelina McMillan-Major, and Shmargaret
  Shmitchell. 2021.
\newblock On the dangers of stochastic parrots: Can language models be too big?
\newblock In \emph{Proceedings of the 2021 ACM Conference on Fairness,
  Accountability, and Transparency}, pages 610--623.

\bibitem[{Bogost(2022)}]{bogost.chatgpt_incorrect}
Ian Bogost. 2022.
\newblock \href
  {https://www.theatlantic.com/technology/archive/2022/12/chatgpt-openai-artificial-intelligence-writing-ethics/672386/}
  {{ChatGPT} is dumber than you think}.
\newblock \emph{The Atlantic}.

\bibitem[{Bowman et~al.(2015)Bowman, Angeli, Potts, and Manning}]{SNLI}
{Samuel R.} Bowman, Gabor Angeli, Christopher Potts, and {Christopher D.}
  Manning. 2015.
\newblock \href {https://doi.org/10.18653/v1/d15-1075} {A large annotated
  corpus for learning natural language inference}.
\newblock In \emph{Proceedings of the 2015 Conference on Empirical Methods in
  Natural Language Processing (EMNLP)}, pages 632--642, Lisbon, Portugal.
  Association for Computational Linguistics.

\bibitem[{Brown et~al.(2020)Brown, Mann, Ryder, Subbiah, Kaplan, Dhariwal,
  Neelakantan, Shyam, Sastry, Askell, Agarwal, Herbert-Voss, Krueger, Henighan,
  Child, Ramesh, Ziegler, Wu, Winter, Hesse, Chen, Sigler, Litwin, Gray, Chess,
  Clark, Berner, McCandlish, Radford, Sutskever, and Amodei}]{GPT3}
Tom Brown, Benjamin Mann, Nick Ryder, Melanie Subbiah, Jared~D. Kaplan,
  Prafulla Dhariwal, Arvind Neelakantan, Pranav Shyam, Girish Sastry, Amanda
  Askell, Sandhini Agarwal, Ariel Herbert-Voss, Gretchen Krueger, Tom Henighan,
  Rewon Child, Aditya Ramesh, Daniel Ziegler, Jeffrey Wu, Clemens Winter, Chris
  Hesse, Mark Chen, Eric Sigler, Mateusz Litwin, Scott Gray, Benjamin Chess,
  Jack Clark, Christopher Berner, Sam McCandlish, Alec Radford, Ilya Sutskever,
  and Dario Amodei. 2020.
\newblock \href
  {https://proceedings.neurips.cc/paper/2020/file/1457c0d6bfcb4967418bfb8ac142f64a-Paper.pdf}
  {Language models are few-shot learners}.
\newblock In \emph{Advances in Neural Information Processing Systems},
  volume~33, pages 1877--1901.

\bibitem[{Camburu et~al.(2020)Camburu, Shillingford, Minervini, Lukasiewicz,
  and Blunsom}]{camburu2020make}
Oana-Maria Camburu, Brendan Shillingford, Pasquale Minervini, Thomas
  Lukasiewicz, and Phil Blunsom. 2020.
\newblock \href {https://doi.org/10.18653/v1/2020.acl-main.382} {Make up your
  mind! {A}dversarial generation of inconsistent natural language
  explanations}.
\newblock In \emph{Proceedings of the 58th Annual Meeting of the Association
  for Computational Linguistics}, pages 4157--4165, Online. Association for
  Computational Linguistics.

\bibitem[{Candela-Quinonero et~al.(2006)Candela-Quinonero, Dagan, Magnini, and
  d'Alch{\'e} Buc}]{RTE}
Joaquin Candela-Quinonero, Ido Dagan, Bernardo Magnini, and Florence
  d'Alch{\'e} Buc. 2006.
\newblock Evaluating {P}redictive {U}ncertainty, {V}isual {O}bjects
  {C}lassification and {R}ecognising {T}extual {E}ntailment: {S}elected
  {P}roceedings of the {F}irst {PASCAL} {M}achine {L}earning {C}hallenges
  {W}orkshop.

\bibitem[{Choi et~al.(2023)Choi, Hickman, Monahan, and
  Schwarcz}]{choi2023chatgpt}
Jonathan~H. Choi, Kristin~E. Hickman, Amy Monahan, and Daniel~B. Schwarcz.
  2023.
\newblock \href {https://ssrn.com/abstract=4335905} {{ChatGPT} goes to law
  school}.
\newblock \emph{Minnesota Legal Studies Research Paper}.

\bibitem[{Clark et~al.(2020)Clark, Luong, Le, and Manning}]{electra}
Kevin Clark, Minh-Thang Luong, Quoc~V. Le, and Christopher~D. Manning. 2020.
\newblock \href {https://openreview.net/forum?id=r1xMH1BtvB} {{ELECTRA:
  P}re-training text encoders as discriminators rather than generators}.
\newblock In \emph{Proceedings of the International Conference on Learning
  Representations}.

\bibitem[{De~Visser et~al.(2016)De~Visser, Monfort, McKendrick, Smith,
  McKnight, Krueger, and Parasuraman}]{de2016almost}
Ewart~J. De~Visser, Samuel~S. Monfort, Ryan McKendrick, Melissa A.~B. Smith,
  Patrick~E. McKnight, Frank Krueger, and Raja Parasuraman. 2016.
\newblock \href {https://psycnet.apa.org/record/2016-37452-001} {Almost human:
  Anthropomorphism increases trust resilience in cognitive agents.}
\newblock \emph{Journal of Experimental Psychology: Applied}, 22(3):331.

\bibitem[{Devlin et~al.(2019)Devlin, Chang, Lee, and Toutanova}]{BERT}
Jacob Devlin, Ming-Wei Chang, Kenton Lee, and Kristina Toutanova. 2019.
\newblock \href {https://doi.org/10.18653/v1/N19-1423} {{BERT}: Pre-training of
  deep bidirectional transformers for language understanding}.
\newblock In \emph{Proceedings of the 2019 Conference of the North {A}merican
  Chapter of the Association for Computational Linguistics (NAACL): Human
  Language Technologies, Volume 1 (Long and Short Papers)}, pages 4171--4186.

\bibitem[{Dolan and Brockett(2005)}]{MRPC}
William~B. Dolan and Chris Brockett. 2005.
\newblock Automatically constructing a corpus of sentential paraphrases.
\newblock In \emph{Proceedings of the Third International Workshop on
  Paraphrasing (IWP2005)}.

\bibitem[{Elazar et~al.(2021)Elazar, Kassner, Ravfogel, Ravichander, Hovy,
  Sch{\"u}tze, and Goldberg}]{elazar2021erratum}
Yanai Elazar, Nora Kassner, Shauli Ravfogel, Abhilasha Ravichander, Eduard
  Hovy, Hinrich Sch{\"u}tze, and Yoav Goldberg. 2021.
\newblock \href {https://doi.org/10.1162/tacl_x_00455} {Erratum: Measuring and
  improving consistency in pretrained language models}.
\newblock \emph{Transactions of the Association for Computational Linguistics},
  9:1407--1407.

\bibitem[{Ettinger(2020)}]{ettinger2020bert}
Allyson Ettinger. 2020.
\newblock What {BERT} is not: Lessons from a new suite of psycholinguistic
  diagnostics for language models.
\newblock \emph{Transactions of the Association for Computational Linguistics},
  8:34--48.

\bibitem[{Fares(2023)}]{fares.chatgpt}
Omar~H. Fares. 2023.
\newblock \href
  {https://theconversation.com/chatgpt-could-be-a-game-changer-for-marketers-but-it-wont-replace-humans-any-time-soon-198053}
  {{ChatGPT} could be a game-changer for marketers, but it won’t replace
  humans any time soon}.
\newblock \emph{The Conversation}.

\bibitem[{Frieder et~al.(2023)Frieder, Pinchetti, Griffiths, Salvatori,
  Lukasiewicz, Petersen, Chevalier, and Berner}]{frieder.mathmathics}
Simon Frieder, Luca Pinchetti, Ryan-Rhys Griffiths, Tommaso Salvatori, Thomas
  Lukasiewicz, Philipp~Christian Petersen, Alexis Chevalier, and Julius Berner.
  2023.
\newblock \href {https://doi.org/10.48550/ARXIV.2301.13867} {Mathematical
  capabilities of {ChatGPT}}.

\bibitem[{Gazes et~al.(2012)Gazes, Chee, and Hampton}]{gazes2012cognitive}
Regina~Paxton Gazes, Nicholas~W. Chee, and Robert~R. Hampton. 2012.
\newblock Cognitive mechanisms for transitive inference performance in rhesus
  monkeys: Measuring the influence of associative strength and inferred order.
\newblock \emph{Journal of Experimental Psychology: Animal Behavior Processes},
  38(4):331.

\bibitem[{Hossain et~al.(2020)Hossain, Kovatchev, Dutta, Kao, Wei, and
  Blanco}]{hossain2020analysis}
Md~Mosharaf Hossain, Venelin Kovatchev, Pranoy Dutta, Tiffany Kao, Elizabeth
  Wei, and Eduardo Blanco. 2020.
\newblock \href {https://doi.org/10.18653/v1/2020.emnlp-main.732} {An analysis
  of natural language inference benchmarks through the lens of negation}.
\newblock In \emph{Proceedings of the 2020 Conference on Empirical Methods in
  Natural Language Processing (EMNLP)}, pages 9106--9118, Online. Association
  for Computational Linguistics.

\bibitem[{Hosseini et~al.(2021)Hosseini, Reddy, Bahdanau, Hjelm, Sordoni, and
  Courville}]{hosseini2021understanding}
Arian Hosseini, Siva Reddy, Dzmitry Bahdanau, R~Devon Hjelm, Alessandro
  Sordoni, and Aaron Courville. 2021.
\newblock \href {https://doi.org/10.18653/v1/2021.naacl-main.102}
  {Understanding by understanding not: Modeling negation in language models}.
\newblock In \emph{Proceedings of the 2021 Conference of the North American
  Chapter of the Association for Computational Linguistics: Human Language
  Technologies}, pages 1301--1312, Online. Association for Computational
  Linguistics.

\bibitem[{Huang et~al.(2020)Huang, Kroening, Ruan, Sharp, Sun, Thamo, Wu, and
  Yi}]{huang2020survey}
Xiaowei Huang, Daniel Kroening, Wenjie Ruan, James Sharp, Youcheng Sun, Emese
  Thamo, Min Wu, and Xinping Yi. 2020.
\newblock A survey of safety and trustworthiness of deep neural networks:
  Verification, testing, adversarial attack and defence, and interpretability.
\newblock \emph{Computer Science Review}, 37:100270.

\bibitem[{Jang et~al.(2022{\natexlab{a}})Jang, Kwon, and
  Lukasiewicz}]{jang2022becel}
Myeongjun Jang, Deuk~Sin Kwon, and Thomas Lukasiewicz. 2022{\natexlab{a}}.
\newblock \href {https://aclanthology.org/2022.coling-1.324} {{BECEL}:
  Benchmark for consistency evaluation of language models}.
\newblock In \emph{Proceedings of the 29th International Conference on
  Computational Linguistics}, pages 3680--3696, Gyeongju, Republic of Korea.
  International Committee on Computational Linguistics.

\bibitem[{Jang et~al.(2023)Jang, Majumder, McAuley, Lukasiewicz, and
  Camburu}]{jang2023know}
Myeongjun Jang, Bodhisattwa~Prasad Majumder, Julian McAuley, Thomas
  Lukasiewicz, and Oana-Maria Camburu. 2023.
\newblock \href {https://doi.org/10.18653/v1/2023.acl-short.47} {{KNOW} how to
  make up your mind! adversarially detecting and alleviating inconsistencies in
  natural language explanations}.
\newblock In \emph{Proceedings of the 61st Annual Meeting of the Association
  for Computational Linguistics (Volume 2: Short Papers)}, pages 540--553,
  Toronto, Canada. Association for Computational Linguistics.

\bibitem[{Jang et~al.(2022{\natexlab{b}})Jang, Mtumbuka, and
  Lukasiewicz}]{jang2022beyond}
Myeongjun Jang, Frank Mtumbuka, and Thomas Lukasiewicz. 2022{\natexlab{b}}.
\newblock \href {https://doi.org/10.18653/v1/2022.findings-naacl.156} {Beyond
  distributional hypothesis: Let language models learn meaning-text
  correspondence}.
\newblock In \emph{Findings of the Association for Computational Linguistics:
  NAACL 2022}, pages 2030--2042, Seattle, United States. Association for
  Computational Linguistics.

\bibitem[{Jiang et~al.(2020)Jiang, Xu, Araki, and Neubig}]{jiang2020can}
Zhengbao Jiang, Frank~F. Xu, Jun Araki, and Graham Neubig. 2020.
\newblock How can we know what language models know?
\newblock \emph{Transactions of the Association for Computational Linguistics},
  8:423--438.

\bibitem[{Jung et~al.(2019)Jung, Dong, and Lee}]{jung2019neural}
Eun-Soo Jung, Suh-Yeon Dong, and Soo-Young Lee. 2019.
\newblock \href {https://www.nature.com/articles/s41598-019-46098-8} {Neural
  correlates of variations in human trust in human-like machines during
  non-reciprocal interactions}.
\newblock \emph{Scientific Reports}, 9(1):1--10.

\bibitem[{Jung et~al.(2022)Jung, Qin, Welleck, Brahman, Bhagavatula, Le~Bras,
  and Choi}]{jung-etal-2022-maieutic}
Jaehun Jung, Lianhui Qin, Sean Welleck, Faeze Brahman, Chandra Bhagavatula,
  Ronan Le~Bras, and Yejin Choi. 2022.
\newblock \href {https://doi.org/10.18653/v1/2022.emnlp-main.82} {Maieutic
  prompting: Logically consistent reasoning with recursive explanations}.
\newblock In \emph{Proceedings of the 2022 Conference on Empirical Methods in
  Natural Language Processing}, pages 1266--1279, Abu Dhabi, United Arab
  Emirates. Association for Computational Linguistics.

\bibitem[{Kassner and Sch{\"u}tze(2020)}]{kassner2020negated}
Nora Kassner and Hinrich Sch{\"u}tze. 2020.
\newblock \href {https://doi.org/10.18653/v1/2020.acl-main.698} {Negated and
  misprimed probes for pretrained language models: Birds can talk, but cannot
  fly}.
\newblock In \emph{Proceedings of the 58th Annual Meeting of the Association
  for Computational Linguistics}, pages 7811--7818, Online. Association for
  Computational Linguistics.

\bibitem[{Kim et~al.(2021)Kim, Kim, Park, and Kang}]{kim2021learn}
Gangwoo Kim, Hyunjae Kim, Jungsoo Park, and Jaewoo Kang. 2021.
\newblock \href {https://doi.org/10.18653/v1/2021.acl-long.478} {Learn to
  resolve conversational dependency: A consistency training framework for
  conversational question answering}.
\newblock In \emph{Proceedings of the 59th Annual Meeting of the Association
  for Computational Linguistics and the 11th International Joint Conference on
  Natural Language Processing (Volume 1: Long Papers)}, pages 6130--6141,
  Online. Association for Computational Linguistics.

\bibitem[{Kumar and Joshi(2022)}]{kumar-joshi-2022-striking}
Ashutosh Kumar and Aditya Joshi. 2022.
\newblock \href {https://doi.org/10.18653/v1/2022.findings-acl.148} {Striking a
  balance: Alleviating inconsistency in pre-trained models for symmetric
  classification tasks}.
\newblock In \emph{Findings of the Association for Computational Linguistics:
  ACL 2022}, pages 1887--1895, Dublin, Ireland. Association for Computational
  Linguistics.

\bibitem[{Kung et~al.(2023)Kung, Cheatham, Medenilla, Sillos, De~Leon,
  Elepa{\~n}o, Madriaga, Aggabao, Diaz-Candido, Maningo
  et~al.}]{kung2023performance}
Tiffany~H. Kung, Morgan Cheatham, Arielle Medenilla, Czarina Sillos, Lorie
  De~Leon, Camille Elepa{\~n}o, Maria Madriaga, Rimel Aggabao, Giezel
  Diaz-Candido, James Maningo, et~al. 2023.
\newblock Performance of {ChatGPT on USMLE: P}otential for {AI}-assisted
  medical education using large language models.
\newblock \emph{PLOS Digital Health}, 2(2):e0000198.

\bibitem[{Lester et~al.(2021)Lester, Al-Rfou, and Constant}]{lester2021power}
Brian Lester, Rami Al-Rfou, and Noah Constant. 2021.
\newblock \href {https://doi.org/10.18653/v1/2021.emnlp-main.243} {The power of
  scale for parameter-efficient prompt tuning}.
\newblock In \emph{Proceedings of the 2021 Conference on Empirical Methods in
  Natural Language Processing}, pages 3045--3059, Online and Punta Cana,
  Dominican Republic. Association for Computational Linguistics.

\bibitem[{Li et~al.(2019)Li, Gupta, Mehta, and Srikumar}]{li2019logic}
Tao Li, Vivek Gupta, Maitrey Mehta, and Vivek Srikumar. 2019.
\newblock \href {https://doi.org/10.18653/v1/D19-1405} {A logic-driven
  framework for consistency of neural models}.
\newblock In \emph{Proceedings of the 2019 Conference on Empirical Methods in
  Natural Language Processing and the 9th International Joint Conference on
  Natural Language Processing (EMNLP-IJCNLP)}, pages 3924--3935, Hong Kong,
  China. Association for Computational Linguistics.

\bibitem[{Lin and Ng(2022)}]{lin-ng-2022-bert}
Ruixi Lin and Hwee~Tou Ng. 2022.
\newblock \href {https://doi.org/10.18653/v1/2022.acl-short.11} {Does {BERT}
  know that the {IS}-a relation is transitive?}
\newblock In \emph{Proceedings of the 60th Annual Meeting of the Association
  for Computational Linguistics (Volume 2: Short Papers)}, pages 94--99,
  Dublin, Ireland. Association for Computational Linguistics.

\bibitem[{Milmo(2023)}]{milmo.chatgpt_100m}
Dan Milmo. 2023.
\newblock \href
  {https://www.theguardian.com/technology/2023/feb/02/chatgpt-100-million-users-open-ai-fastest-growing-app}
  {{ChatGPT} reaches 100 million users two months after launch}.
\newblock \emph{The Guardian}.

\bibitem[{Mitchell et~al.(2022)Mitchell, Noh, Li, Armstrong, Agarwal, Liu,
  Finn, and Manning}]{mitchell2022enhancing}
Eric Mitchell, Joseph~J. Noh, Siyan Li, William~S. Armstrong, Ananth Agarwal,
  Patrick Liu, Chelsea Finn, and Christopher~D. Manning. 2022.
\newblock \href {https://ericmitchell.ai/concord.pdf} {Enhancing
  self-consistency and performance of pretrained language models with {NLI}}.
\newblock In \emph{Proceedings of the 2022 Conference on Empirical Methods in
  Natural Language Processing (EMNLP)}. Association for Computational
  Linguistics.

\bibitem[{Ouyang et~al.(2022)Ouyang, Wu, Jiang, Almeida, Wainwright, Mishkin,
  Zhang, Agarwal, Slama, Ray et~al.}]{ouyang2022training}
Long Ouyang, Jeffrey Wu, Xu~Jiang, Diogo Almeida, Carroll Wainwright, Pamela
  Mishkin, Chong Zhang, Sandhini Agarwal, Katarina Slama, Alex Ray, et~al.
  2022.
\newblock Training language models to follow instructions with human feedback.
\newblock In \emph{Advances in Neural Information Processing Systems},
  volume~35, pages 27730--27744.

\bibitem[{Pilehvar and Camacho-Collados(2019)}]{WiC}
Mohammad~Taher Pilehvar and Jose Camacho-Collados. 2019.
\newblock \href {https://doi.org/10.18653/v1/N19-1128} {{W}i{C}: the
  word-in-context dataset for evaluating context-sensitive meaning
  representations}.
\newblock In \emph{Proceedings of the 2019 Conference of the North {A}merican
  Chapter of the Association for Computational Linguistics: Human Language
  Technologies, Volume 1 (Long and Short Papers)}, pages 1267--1273,
  Minneapolis, Minnesota. Association for Computational Linguistics.

\bibitem[{Raffel et~al.(2020)Raffel, Shazeer, Roberts, Lee, Narang, Matena,
  Zhou, Li, and Liu}]{T5}
Colin Raffel, Noam Shazeer, Adam Roberts, Katherine Lee, Sharan Narang, Michael
  Matena, Yanqi Zhou, Wei Li, and Peter~J. Liu. 2020.
\newblock \href {https://w.jmlr.org/papers/volume21/20-074/20-074.pdf}
  {Exploring the limits of transfer learning with a unified text-to-text
  transformer}.
\newblock \emph{Journal of Machine Learning Research}, 21:1--67.

\bibitem[{Raj et~al.(2022)Raj, Rosati, and Majumdar}]{raj2022measuring}
Harsh Raj, Domenic Rosati, and Subhabrata Majumdar. 2022.
\newblock \href {https://openreview.net/pdf?id=SgbpddeEV-C} {Measuring
  reliability of large language models through semantic consistency}.
\newblock In \emph{NeurIPS 2022 Workshop on Machine Learning Safety}.

\bibitem[{Ravichander et~al.(2020)Ravichander, Hovy, Suleman, Trischler, and
  Cheung}]{ravichander2020systematicity}
Abhilasha Ravichander, Eduard Hovy, Kaheer Suleman, Adam Trischler, and Jackie
  Chi~Kit Cheung. 2020.
\newblock \href {https://aclanthology.org/2020.starsem-1.10} {On the
  systematicity of probing contextualized word representations: The case of
  hypernymy in {BERT}}.
\newblock In \emph{Proceedings of the Ninth Joint Conference on Lexical and
  Computational Semantics}, pages 88--102, Barcelona, Spain (Online).
  Association for Computational Linguistics.

\bibitem[{Skopeliti and Milmo(2023)}]{skopeliti.chatgpt_incorrect_facts}
Clea Skopeliti and Dan Milmo. 2023.
\newblock \href
  {https://www.theguardian.com/technology/2023/feb/08/chatgpt-users-views-ai-chatbot-essays-emails}
  {‘{ChatGPT} needs a huge amount of editing’: users’ views mixed on {AI}
  chatbot}.
\newblock \emph{The Guardian}.

\bibitem[{Strubell et~al.(2019)Strubell, Ganesh, and
  McCallum}]{strubell2019energy}
Emma Strubell, Ananya Ganesh, and Andrew McCallum. 2019.
\newblock \href {https://doi.org/10.18653/v1/P19-1355} {Energy and policy
  considerations for deep learning in {NLP}}.
\newblock In \emph{Proceedings of the 57th Annual Meeting of the Association
  for Computational Linguistics}, pages 3645--3650, Florence, Italy.
  Association for Computational Linguistics.

\bibitem[{Tam et~al.(2023)Tam, Mascarenhas, Zhang, Kwan, Bansal, and
  Raffel}]{tam2022evaluating}
Derek Tam, Anisha Mascarenhas, Shiyue Zhang, Sarah Kwan, Mohit Bansal, and
  Colin Raffel. 2023.
\newblock \href {https://doi.org/10.18653/v1/2023.findings-acl.322} {Evaluating
  the factual consistency of large language models through news summarization}.
\newblock In \emph{Findings of the Association for Computational Linguistics:
  ACL 2023}, pages 5220--5255, Toronto, Canada. Association for Computational
  Linguistics.

\bibitem[{Terwiesch(2023)}]{terwiesch2023would}
Christian Terwiesch. 2023.
\newblock Would {ChatGPT} get a {Wharton MBA? A} prediction based on its
  performance in the operations management course.
\newblock \emph{Mack Institute for Innovation Management at the Wharton School,
  University of Pennsylvania. Retrieved from: https://mackinstitute. wharton.
  upenn. edu/wpcontent/uploads/2023/01/Christian-Terwiesch-Chat-GTP-1.24. pdf
  [Date accessed: February 6th, 2023]}.

\bibitem[{Wang et~al.(2019)Wang, Sun, and Xing}]{wang2019if}
Haohan Wang, Da~Sun, and Eric~P. Xing. 2019.
\newblock What if we simply swap the two text fragments? {A} straightforward
  yet effective way to test the robustness of methods to confounding signals in
  nature language inference tasks.
\newblock In \emph{Proceedings of the AAAI Conference on Artificial
  Intelligence}, volume~33, pages 7136--7143.

\bibitem[{Wang et~al.(2023)Wang, Wang, Li, Gao, Yin, and
  Ren}]{wang-etal-2023-scott}
Peifeng Wang, Zhengyang Wang, Zheng Li, Yifan Gao, Bing Yin, and Xiang Ren.
  2023.
\newblock \href {https://doi.org/10.18653/v1/2023.acl-long.304} {{SCOTT}:
  Self-consistent chain-of-thought distillation}.
\newblock In \emph{Proceedings of the 61st Annual Meeting of the Association
  for Computational Linguistics (Volume 1: Long Papers)}, pages 5546--5558,
  Toronto, Canada. Association for Computational Linguistics.

\bibitem[{Wang and Henao(2021)}]{wang2021unsupervised}
Rui Wang and Ricardo Henao. 2021.
\newblock \href {https://doi.org/10.18653/v1/2021.emnlp-main.430} {Unsupervised
  paraphrasing consistency training for low resource named entity recognition}.
\newblock In \emph{Proceedings of the 2021 Conference on Empirical Methods in
  Natural Language Processing}, pages 5303--5308, Online and Punta Cana,
  Dominican Republic. Association for Computational Linguistics.

\bibitem[{Wei et~al.(2022)Wei, Bosma, Zhao, Guu, Yu, Lester, Du, Dai, and
  Le}]{wei2022finetuned}
Jason Wei, Maarten Bosma, Vincent Zhao, Kelvin Guu, Adams~Wei Yu, Brian Lester,
  Nan Du, Andrew~M Dai, and Quoc~V Le. 2022.
\newblock \href {https://openreview.net/forum?id=gEZrGCozdqR} {Finetuned
  language models are zero-shot learners}.
\newblock In \emph{International Conference on Learning Representations}.

\bibitem[{Wen and Wang(2023)}]{wen2023future}
Jun Wen and Wei Wang. 2023.
\newblock The future of {ChatGPT} in academic research and publishing: A
  commentary for clinical and translational medicine.
\newblock \emph{Clinical and Translational Medicine}, 13(3):e1207--e1207.

\bibitem[{Zhao et~al.(2023)Zhao, Li, Chia, Ding, and Bing}]{zhao2023can}
Ruochen Zhao, Xingxuan Li, Yew~Ken Chia, Bosheng Ding, and Lidong Bing. 2023.
\newblock Can {ChatGPT}-like generative models guarantee factual accuracy? on
  the mistakes of new generation search engines.
\newblock \emph{arXiv preprint arXiv:2304.11076}.

\bibitem[{Zheng et~al.(2021)Zheng, Dong, Huang, Wang, Chi, Singhal, Che, Liu,
  Song, and Wei}]{zheng2021consistency}
Bo~Zheng, Li~Dong, Shaohan Huang, Wenhui Wang, Zewen Chi, Saksham Singhal,
  Wanxiang Che, Ting Liu, Xia Song, and Furu Wei. 2021.
\newblock \href {https://doi.org/10.18653/v1/2021.acl-long.264} {Consistency
  regularization for cross-lingual fine-tuning}.
\newblock In \emph{Proceedings of the 59th Annual Meeting of the Association
  for Computational Linguistics and the 11th International Joint Conference on
  Natural Language Processing (Volume 1: Long Papers)}, pages 3403--3417,
  Online. Association for Computational Linguistics.

\bibitem[{Zhuo et~al.(2023)Zhuo, Huang, Chen, and Xing}]{zhuo2023red}
Terry~Yue Zhuo, Yujin Huang, Chunyang Chen, and Zhenchang Xing. 2023.
\newblock \href {http://arxiv.org/abs/2301.12867} {Red teaming chatgpt via
  jailbreaking: Bias, robustness, reliability and toxicity}.

\end{thebibliography}
\bibliographystyle{acl_natbib}

\begin{acronym}
    \acro{PLM}{pre-trained language model}
    \acro{CRM}{conceptual role model}
    \acro{LM}{language model}
    \acro{LLM}{large language model}
    \acro{NLP}{natural language processing}
    \acro{NLG}{natural language generation}
    \acro{NLU}{natural language understanding}
    \acro{AI}{artificial intelligence}
    \acro{NLI}{natural language inference}
    \acro{QA}{question answering}
    \acro{MRC}{machine reading comprehension}
    \acro{TC}{topic classification}
    \acro{SA}{semantic analysis}
    \acro{WiC}{words-in-context}
    \acro{STS}{semantic textual similarity}
    \acro{MTT}{meaning-text theory}
    \acro{MLM}{masked language modelling}
    \acro{E}{entailment}
    \acro{N}{neutral}
    \acro{C}{contradiction}
\end{acronym}

\clearpage
\onecolumn
\appendix

\section{Prompt Design}~\label{section:prompt_design}

\begin{table*}[ht]
	\begin{center}
		\renewcommand{\arraystretch}{1.1}
		\footnotesize{
			\centering{\setlength\tabcolsep{2.4pt}
		\begin{tabular}{p{0.08\linewidth} | p{0.9\linewidth}}
		\toprule
		\multicolumn{2}{c}{\textbf{SNLI}}  \\ \midrule
        Format & \{\textit{s1}\} Question: \{s2\} True, False or Neither? Answer: \\

        Example & A land rover is being driven across a river. Question: A Land Rover is splashing water as it crosses a river. True, False, or Neither? Answer: \\ \hline
        
        \multicolumn{2}{c}{\textbf{RTE}}  \\ \midrule
        Format & \makecell[tl]{\{\textit{s1}\} Question: \{s2\} True or False? Answer:} \\

        Example & The harvest of sea-weeds is not allowed in the Puget Sound because of marine vegetation's vital role in providing habitat to important species. Question: Marine vegetation is  harvested. True or False? Answer: \\ \hline

        \multicolumn{2}{c}{\textbf{MRPC}}  \\ \midrule
        Format & Sentence 1: \{\textit{s1}\} Sentence 2: \{s2\} Question: Do both sentences mean the same thing? Answer: \\

        Example & Sentence 1: The increase reflects lower credit losses and favorable interest rates.	Sentence 2: The gain came as a result of fewer credit losses and lower interest rates. Question: Do both sentences mean the same thing? Answer: \\ \hline

        \multicolumn{2}{c}{\textbf{WiC}}  \\ \midrule
        Format & Sentence 1: \{\textit{s1}\} Sentence 2: \{s2\} Question: Is the word \{\textit{word}\} used in the same way in the two sentences above? Answer: \\

        Example & Sentence 1: It was the deliberation of his act that was insulting.	Sentence 2: It was the deliberation of his act that was insulting.	The deliberations of the jury. Question: Is the word deliberation used in the same way in the two sentences above? Answer: \\

        \bottomrule
        \end{tabular}}}
    \vspace*{-1ex}
	\caption{Eleuther AI prompt formats and their example used in our experiments for each downstream task.} 
	\label{table.prompt_example}%
	\end{center}
    \vspace*{-2ex}

\end{table*}

\begin{table*}[ht]
	\begin{center}
		\renewcommand{\arraystretch}{1.1}
		\footnotesize{
			\centering{\setlength\tabcolsep{2.4pt}
		\begin{tabular}{p{0.08\linewidth} | p{0.9\linewidth}}
		\toprule
		\multicolumn{2}{c}{\textbf{SNLI}}  \\ \midrule
        Format & If \{\textit{s1}\}, can we conclude \{s2\}? \textbackslash n Yes, It's impossible to say, No \textbackslash n Answer: \\

        Example & If A land rover is being driven across a river, can we conclude A Land Rover is splashing water as it crosses a river? \textbackslash n Yes, It's impossible to say, No \textbackslash n Answer: \\ \hline
        
        \multicolumn{2}{c}{\textbf{RTE}}  \\ \midrule
        Format & \makecell[tl]{\{\textit{s1}\} \textbackslash n Based on the paragraph above can we conclude that \{s2\}? Yes, No \textbackslash n Answer:} \\

        Example & The harvest of sea-weeds is not allowed in the Puget Sound because of marine vegetation's vital role in providing habitat to important species. \textbackslash n Based on the paragraph above can we conclude that Marine vegetation is  harvested? \textbackslash n Yes, No \textbackslash n Answer: \\ \hline

        \multicolumn{2}{c}{\textbf{MRPC}}  \\ \midrule
        Format & Here are two sentences: \textbackslash n \{\textit{s1}\} \textbackslash n \{s2\} \textbackslash n Do they have the same meaning? \textbackslash n Yes, No \textbackslash n Answer: \\

        Example & Here are two sentences: \textbackslash n The increase reflects lower credit losses and favorable interest rates. \textbackslash n The gain came as a result of fewer credit losses and lower interest rates. \textbackslash n Do they have the same meaning? \textbackslash n Yes, No \textbackslash n Answer: \\ \hline

        \multicolumn{2}{c}{\textbf{WiC}}  \\ \midrule
        Format & In these two sentences (1) \{\textit{s1}\} (2) \{s2\} , does the word \{\textit{word}\} mean the same thing? \textbackslash n Yes, No \textbackslash n Answer: \\

        Example & In these two sentences (1) It was the deliberation of his act that was insulting.	(2) It was the deliberation of his act that was insulting, does the word deliberation mean the same thing? \textbackslash n Yes, No \textbackslash n Answer: \\

        \bottomrule
        \end{tabular}}}
    \vspace*{-1ex}
	\caption{Prompt formats designed by \citet{wei2022finetuned} and their examples used in our experiments for each downstream task.} 
	\label{table.prompt_example_wei}%
	\end{center}
    \vspace*{-2ex}

\end{table*}

\clearpage
\section{Examples}~\label{section.example}

\begin{table*}[ht]
	\begin{center}
		\renewcommand{\arraystretch}{1.0}
		\footnotesize{
			\centering{\setlength\tabcolsep{3pt}
    		      \begin{tabular}{p{0.48\linewidth} | p{0.48\linewidth}}
					\toprule
                    \multicolumn{2}{c}{\makecell{\textsc{Task}: RTE,  \textsc{Paraphrase Type}: BECEL}} \\
                    
                    \makecell[c]{\textsc{Original Inputs}} & \makecell[c]{\textsc{Perturbed Inputs}} \\
                    
                    \textsc{S1}: Note that SBB, CFF and FFS stand out for the main railway company, in German, French and Italian. & 
                    \textsc{S1}: Note that SBB, CFF and FFS stand out for the main railway company, in German, French and Italian. \\
                    
                    \textsc{S2}: The French railway company is called SNCF. & 
                    \textsc{S2}: SNCF is the French railway company. \\
                    
                    \makecell[l]{\textsc{Prediction}: Not Entailment} & 
                    \makecell[l]{\textsc{Prediction}: Entailment} \\
                    \midrule
                    
                    \multicolumn{2}{c}{\makecell{\textsc{Task}: SNLI, \textsc{Paraphrase Type}: ChatGPT}} \\
                    
                    \makecell[c]{\textsc{Original Inputs}} & \makecell[c]{\textsc{Perturbed Inputs}} \\
                    
                    \textsc{Premise}: A person swimming in a swimming pool. & 
                    \textsc{Premise}: A person swimming in a swimming pool. \\
                    
                   \textsc{Hypothesis}: A person enjoying the waters. & 
                    \textsc{Hypothesis}: An individual is relishing the water. \\
                    
                    \makecell[l]{\textsc{Prediction}: Not Entailment (Neutral)} & 
                    \makecell[l]{\textsc{Prediction}: Entailment} \\ \midrule

                    \multicolumn{2}{c}{\makecell{\textsc{Task}: WiC, \textsc{Paraphrase Type}: BECEL}} \\
                    
                    \makecell[c]{\textsc{Original Inputs}} & \makecell[c]{\textsc{Perturbed Inputs}} \\

                    \textsc{Word}: glaze & \textsc{Word}: glaze \\
                    \textsc{S1}: Glaze the bread with eggwhite. & 
                    \textsc{S1}: Glaze the bread with eggwhite. \\
                    
                   \textsc{S2}: The potter glazed the dishes. & 
                    \textsc{S2}: The dishes were glazed by the potter. \\
                    
                    \makecell[l]{\textsc{Prediction}: Equivalent} & 
                    \makecell[l]{\textsc{Prediction}: Not Equivalent} \\ \bottomrule
                    
		\end{tabular}}}
	\vspace{-1ex}
	\caption{More examples of semantic consistency violation.} \label{table.more_semantic_example}%
	\end{center}
    \vspace{-2ex}
\end{table*}

\begin{table*}[ht]
	\begin{center}
		\renewcommand{\arraystretch}{1.0}
		\footnotesize{
			\centering{\setlength\tabcolsep{3pt}
        		\begin{tabular}{p{0.48\linewidth} | p{0.48\linewidth}}
					\toprule
     
                    \multicolumn{2}{c}{\makecell{\textsc{Task}: MRPC,  \textsc{Consistency Type}: Negation}} \\
                    
                    \makecell[c]{\textsc{Original Inputs}} & \makecell[c]{\textsc{Perturbed Inputs}} \\

                    \textsc{S1}: The dead cavalry have been honored for more than a century with a hilltop granite obelisk and white headstones . & 
                    \textsc{S2}: The dead cavalry have been honored for more than a century with a hilltop granite obelisk and white headstones . \\
                    
                    \textsc{S2}: The dead cavalrymen are honored with a hilltop granite obelisk and \underline{white} headstones . & 
                    \textsc{S2}: The dead cavalrymen are honored with a hilltop granite  obelisk and \underline{black} headstones. \\
                    
                    \makecell[l]{\textsc{Prediction}: Equivalent} & 
                    \makecell[l]{\textsc{Prediction}: Equivalent} \\
                    \midrule

                    \multicolumn{2}{c}{\makecell{\textsc{Task}: SNLI,  \textsc{Consistency Type}: Negation}} \\
                    
                    \makecell[c]{\textsc{Original Inputs}} & \makecell[c]{\textsc{Perturbed Inputs}} \\

                    \textsc{Premise}: A young man wearing goggles is putting some liquid in a beaker while a young girl wearing blue gloves looks down while holding a pen. & 
                    \textsc{Premise}: A young man wearing goggles is putting some liquid in a beaker while a young girl wearing blue gloves looks down while holding a pen. \\
                    
                    \textsc{Hypothesis}: Two people are in the same area. & 
                    \textsc{Hypothesis}: Two people are \underline{not} in the same area. \\
                    
                    \makecell[l]{\textsc{Prediction}: Entailment} & 
                    \makecell[l]{\textsc{Prediction}: Entailment} \\
                    \midrule

                    \multicolumn{2}{c}{\makecell{\textsc{Task}: MRPC,  \textsc{Consistency Type}: Symmetric}} \\
                    
                    \makecell[c]{\textsc{Original Inputs}} & \makecell[c]{\textsc{Perturbed Inputs}} \\
                    
                    \textsc{S1}: In 2001 , the diocese reached a \$ 15 million settlement  involving five priests and 26 plaintiffs . & 
                    \textsc{S1}: The diocese reached a settlement in 2001 involving five  priests and 26 plaintiffs for an undisclosed sum . \\
                    
                    \textsc{S2}: The diocese reached a settlement in 2001 involving five  priests and 26 plaintiffs for an undisclosed sum . & 
                    \textsc{S2}: In 2001 , the diocese reached a \$ 15 million settlement  involving five priests and 26 plaintiffs . \\
                    
                    \makecell[l]{\textsc{Prediction}: Not Equivalent} & 
                    \makecell[l]{\textsc{Prediction}: Equivalent} \\ \midrule

                    \multicolumn{2}{c}{\makecell{\textsc{Task}: WiC,  \textsc{Consistency Type}: Symmetric}} \\
                    
                    \makecell[c]{\textsc{Original Inputs}} & \makecell[c]{\textsc{Perturbed Inputs}} \\

                    \textsc{Word}: master & \textsc{Word}: master \\
                    \textsc{S1}: One of the old masters. & 
                    \textsc{S1}: A master of the violin. \\
                    
                    \textsc{S2}: A master of the violin. & 
                    \textsc{S2}: One of the old masters. \\
                    
                    \makecell[l]{\textsc{Prediction}:  Equivalent} & 
                    \makecell[l]{\textsc{Prediction}: Not Equivalent} \\

                    \bottomrule
                    
		\end{tabular}}}
	\vspace{-1ex}
	\caption{More examples of negation and symmetric consistency violations.} \label{table.more_neg_sym_examples}%
	\end{center}
    \vspace{-3ex}
\end{table*}

\begin{table*}[ht]
	\begin{center}
		\renewcommand{\arraystretch}{1.0}
		\footnotesize{
			\centering{\setlength\tabcolsep{3pt}
        		\begin{tabular}{p{0.48\linewidth} | p{0.48\linewidth}}
					\toprule

                    \multicolumn{2}{c}{\makecell{\textsc{Task}: WiC,  \textsc{Consistency Type}: Transitive}} \\

                    \multicolumn{2}{c}{\textsc{Original Inputs}} \\
                    \textsc{Sentence1}: Strike a medal. & 
                    \textsc{Sentence1}: Strike coins. \\
                                        
                    \textsc{Sentence2}: Strike coins. & 
                    \textsc{Sentence2}: A bullet struck him. \\
                    \textsc{Word}: strike & \textsc{Word}: strike \\
                    
                    \textsc{Prediction}: Equivalent  & 
                    \textsc{Prediction}: Not Equivalent \\

                    \multicolumn{2}{c}{\textsc{Newly Created Inputs}} \\
                    \multicolumn{2}{l}{\makecell{\textsc{Sentence1}: Strike a medal \textsc{Sentence2}: A bullet struck him. \textsc{Word}: strike \\                    \textsc{Prediction}: Equivalent
                    }} \\ \midrule

                    \multicolumn{2}{c}{\makecell{\textsc{Task}: SNLI,  \textsc{Consistency Type}: Transitive}} \\

                    \multicolumn{2}{c}{\textsc{Original Inputs}} \\
                    \textsc{Premise}: A performance group is staged in one collective motion. & 
                    \textsc{Premise}: A performance group is staged in one collective motion. \\
                                        
                    \textsc{Hypothesis}: The group is separate from one another. & 
                    \textsc{Hypothesis}: There's a performance group doing something together. \\
                    
                    \textsc{Prediction}: Contradiction  & 
                    \textsc{Prediction}: Entailment \\

                    \multicolumn{2}{c}{\textsc{Newly Created Inputs}} \\
                    \multicolumn{2}{l}{\makecell{\textsc{Premise}: The group is separate from one another. \textsc{Hypothesis}: There's a performance group doing something together. \\                    \textsc{Prediction}: Entailment
                    }} \\ \bottomrule
                    
		\end{tabular}}}
	\vspace{-1ex}
	\caption{More examples of transitive consistency violations.} \label{table.more_transitive_example}%
	\end{center}
    \vspace{-3ex}
\end{table*}

\end{document}